\documentclass{article}
\usepackage[utf8]{inputenc}
\usepackage{amsmath}
\usepackage{amsthm}
\usepackage{amssymb}
\usepackage{graphicx}
\usepackage{subfigure}
\usepackage{ragged2e}
\usepackage{geometry}
\usepackage[font=scriptsize]{caption}
\usepackage{booktabs}
\usepackage{algorithm}
\usepackage{algpseudocode}
\usepackage{algorithm}
\usepackage{algpseudocode}
\usepackage{chngcntr}
\usepackage{mathtools}
\usepackage[english]{babel}
\usepackage{caption}
\usepackage{float}
\usepackage[round]{natbib}
\newcounter{mycount}
\counterwithin{algorithm}{mycount}
\refstepcounter{mycount}

\usepackage{xcolor}
\usepackage{graphicx}

\usepackage[colorlinks = true,
            linkcolor = blue,
            urlcolor  = blue,
            citecolor = blue,
            anchorcolor = blue]{hyperref}
\usepackage{cleveref}
\newtheorem{remark}{Remark}

\newtheorem{theorem}{Theorem}

\newtheorem{assumption}{Assumption}
\newtheorem{proposition}{Proposition}

\newtheorem{innercustomgeneric}{\customgenericname}
\providecommand{\customgenericname}{}
\newcommand{\newcustomtheorem}[2]{%
  \newenvironment{#1}[1]
  {%
   \renewcommand\customgenericname{#2}%
   \renewcommand\theinnercustomgeneric{##1}%
   \innercustomgeneric
  }
  {\endinnercustomgeneric}
}

\newcustomtheorem{customthm}{Theorem}
\newcustomtheorem{customlemma}{Lemma}
\newcustomtheorem{customproposition}{Proposition}
\newcustomtheorem{customdefinition}{Definition}

\usepackage[mathscr]{euscript}
\usepackage[colorlinks = true,
            linkcolor = blue,
            urlcolor  = blue,
            citecolor = blue,
            anchorcolor = blue]{hyperref}

\title{Concentration bounds on response-based vector embeddings of black-box generative models}
\author{Aranyak Acharyya, Joshua Agterberg, Youngser Park, Carey E. Priebe}
\date{\today}

\begin{document}

\maketitle

\begin{abstract}
Generative models, such as large language models or text-to-image diffusion models, can generate relevant responses to user-given queries. Response-based vector embeddings of generative models facilitate statistical analysis and inference on a given collection of black-box generative models. The \textit{Data Kernel Perspective Space embedding} is one particular method of obtaining response-based vector embeddings for a given set of generative models, already discussed in the literature. 
This method obtains pairwise dissimilarities between every pair of generative models in a given set of generative models, based on their responses to a set of user-given queries, and embeds the dissimilarities into a finite-dimensional Euclidean space through multidimensional scaling. In our work, we establish high probability concentration bounds for the resultant vector embeddings based on sampled responses of the generative models. 
\end{abstract}
\section{Introduction}
\label{Sec:Introduction}
Generative models in artificial intelligence have found ubiquitous use in natural language processing
\citep{devlin-etal-2019-bert,NEURIPS2020_1457c0d6,sanh2022multitaskpromptedtrainingenables},
 code generation  \citep{zhang2023planning} and text-to-image generation \citep{crowson2022vqgan}. Large langugae models in particular have the potential to revolutionize human-computer interaction \citep{bubeck2023sparks}. 
With the increase in use of generative models in different spheres of life, there has been a surge in demand for theoretically sound tools which can perform statistical analysis and inference tasks on a collection of black-box generative models. \cite{moniri2024evaluating} proposes a novel method for evaluating the performances of a class of interacting language models. For the purpose of providing performance guarantees and unsupervised learning to investigate differences in model behaviour, various works
\citep{faggioli2023perspectives,duderstadt2023comparing}
explore and demonstrate the potential of response-based vector embeddings of generative models for subsequent inference tasks. 
In particular, \cite{helm2024tracking} discusses a novel technique for obtaining a vector representation for every generative model in a given collection of generative models, based on their responses to a set of queries, which can be used for further downstream inference tasks on the collection of generative models.
\cite{acharyya2024consistent}
investigates the sufficient conditions under which the population-level vector representations for a class of generative models can be estimated consistently from their responses to a set of user-given queries, and \cite{helm2024embedding} investigates the consistency of the sample vector representations for subsequent inference tasks. 
\newline
\newline
In this paper, we establish high probability concentration bounds for the vector representations for a collection of generative models based on their sample responses to a set of queries.
The technique used in this paper for obtaining the vector representations  is a variant of \textit{Data Kernel Perspective Space embedding}, a technique for 
obtaining response-based vector embeddings of generative models,
proposed in \cite{helm2024embedding}.
This method computes dissimilarities between pairs of generative models based on their responses to user-given queries, and embeds these dissimilarities into a vector space. In an ideal scenario, we would know the distribution of responses of the generative models to the queries, which would yield population-level vector embeddings. However, in reality, the true response distributions are unknown and hence as a proxy we use the empirical distribution of sampled responses, to obtain sample-level vector embeddings. Our work investigates the concentration of the sample vector embeddings around their population counterparts (upto an orthogonal transformation). 
\newline
\newline
In regards to the mathematical formulation, our work is in succession of a long line of works in literature that explores non-asymptotic properties of scaled eigenvectors of matrices under noise. In \cite{abbe2022lp}, the authors establish the rate of convergence of the leading left singular vector of the noisy matrix to its ground-truth counterpart. A spectral clustering algorithm is proposed in \cite{agterberg2022joint} to address a joint community detection problem in multilayer networks, which studies perturbation bounds on eigenvalues and eigenvectors of matrices. A generalized framework for Principal Component Analysis in presence of heteroskedastic noise is introduced in \cite{zhang2022heteroskedastic}, which studies the effect of perturbation on the singular subspaces. Our work investigates perturbation bounds on the scaled eigenvectors of dissimilarity matrices of growing size, under reasonable regularity assumptions, and we establish sharp bounds.  
\newline
\newline
Our study is based on a realistic scenario where the number of generative models, the number of queries and the number of samples or replicates can grow together. 
Under appropriate regularity conditions, these concentration bounds tell us that in order to estimate the population-level vector representations with a certain accuracy, how many sample responses are needed. 
\newline
\newline
We arrange the manuscript in the following manner. In  \Cref{Subsec:Notations}, we introduce our notations and terminologies. In \Cref{Sec:Background}, we describe the background of our work, and what motivates us to conduct this investigation. Then, we 
describe the setting of our paper in \Cref{Sec:Decsription_of_DKPS}. We state our theoretical results in  \Cref{Sec:Theoretical_results}, which are followed by the findings from numerical experiments in  \Cref{Sec:Numerical_Experiments}. Finally, we discuss the significance of our work in  \Cref{Sec:Discussion}, along with possible future directions. The proofs of our theoretical results are in  \Cref{Sec:Appendix}.
\subsection{Notations}
\label{Subsec:Notations}
In this paper, every vector will be represented by a bold lower case letter such as $\mathbf{v}$. Any vector by default is a column vector. Matrices will be denoted by bold upper case letters such as $\mathbf{A}$. For a matrix $\mathbf{A}$, the $(i,j)$-th entry will be given by $(\mathbf{A})_{i,j}$, the $i$-th row (written as a column vector) will be given by $(\mathbf{A})_{i \cdot}$
and the $j$-th column will be given by $(\mathbf{A})_{\cdot j}$. For any matrix $\mathbf{A} \in \mathbb{R}^{m \times n}$ with $\mathrm{rank}(\mathbf{A})=r$, the singular values in descending order will be given by $\sigma_1(\mathbf{A}) \geq \dots \geq \sigma_r(\mathbf{A})$, the
corresponding left singular vectors will be given by $\mathbf{u}_1(\mathbf{A}),\dots , \mathbf{u}_r(\mathbf{A})$ and the corresponding right singular vectors will be given by $\mathbf{v}_1(\mathbf{A}),\dots \mathbf{v}_r(\mathbf{A})$. The $n \times n$ centering matrix will be denoted by $\mathbf{H}_n=\mathbf{I}_n-
\frac{1}{n}
(\boldsymbol{1}_n \boldsymbol{1}_n^T)$ where $\mathbf{I}_n$ is the $n \times n$ identity matrix and $\boldsymbol{1}_n$ is the $n$-length vector of all ones. For a matrix $\mathbf{A} \in \mathbb{R}^{m \times n}$, $\mathbf{A}_{[p:q,r:s]}$ (where $1 \leq p<q \leq m$, $1 \leq r<s \leq n$) denotes the matrix obtained by retaining the $\left\lbrace p,(p+1),\dots q \right\rbrace$-th rows and $\left\lbrace r,(r+1),\dots s \right\rbrace$-th rows of $\mathbf{A}$, and 
$\mathbf{A}_{[p:q,.]}$ denotes the matrix obtained by retaining the $\left\lbrace p,(p+1),\dots q \right\rbrace$-th rows and all the columns of $\mathbf{A}$, and $\mathbf{A}_{[.,r:s]}$ denotes the matrix obtained by retaining the $\left\lbrace r,(r+1),\dots s \right\rbrace$-th columns and all the rows of $\mathbf{A}$. For two matrices $\mathbf{A},\mathbf{B} \in \mathbb{R}^{m \times n}$,  
the Hadamard product of $\mathbf{A}$ and $\mathbf{B}$ is given by $\mathbf{A} \circ \mathbf{B}$, such that
\begin{equation*}
    (\mathbf{A} \circ \mathbf{B})_{i,j}=
    (
\mathbf{A})_{i,j} (\mathbf{B})_{i,j}.
\end{equation*}
Moreover, for any matrix $\mathbf{A} \in \mathbb{R}^{m \times n}$, and for any $s \in \mathbb{R}$, the Hadamard power $\mathbf{A}^{\circ s} \in \mathbb{R}^{m \times n}$ is such that
\begin{equation*}
    (\mathbf{A}^{\circ s})_{i,j}=
   (
    \mathbf{A})_{ij}^s
    .
\end{equation*}
For any matrix $\mathbf{A}$, 
the spectral norm is denoted by $\left\lVert
\mathbf{A}
\right\rVert$ and
the Frobenius norm is denoted by $\left\lVert \mathbf{A} \right\rVert_F$. 
\newline
\newline
For two sequences $\lbrace a_n \rbrace_{n=1}^{\infty}$ and $\lbrace b_n \rbrace_{n=1}^{\infty}$, 
we use the following notations:
\begin{equation*}
\begin{aligned}
    &a_n=o(b_n) \iff
    b_n=\omega(a_n)
    \hspace{0.5cm}
    \text {  if} \lim_{n \to \infty} \frac{a_n}{b_n}=0, \\
    &a_n=O(b_n) \iff b_n=\Omega(a_n) \hspace{0.5cm} \text{if } \exists \; C>0,n_0 \in \mathbb{N} \text{ such that for all $n \geq n_0$}, \frac{a_n}{b_n} \leq C.
\end{aligned}
\end{equation*} 
\newline
\newline
Discussed below are some important definitions and notions that we will frequently encounter in this paper.
\newline
\section{Background and Motivation}
\label{Sec:Background}
A generative model is a random map from an input space or query space (denoted by $\mathcal{Q}$) to an output space or response space (denoted by $\mathcal{X}$). For every query $q \in \mathcal{Q}$, we denote the response by $f(q)$. 
\newline
\newline
It is of interest to develop methodologies for carrying out statistical inference tasks on a set of generative models. However, generative models are typically black-boxes, that is, for a generative model $f:\mathcal{Q} \to \mathcal{X}$, we typically do not have access to the functional form of $f$. Hence, we
analyse the responses of the generative models to user-given queries to carry out inference on the generative models. We
quantify the response of a generative model to a query as a vector, using
an embedding function $g:\mathcal{X} \to \mathbb{R}^p$, such that the vectorized response $g(f(q)) \in \mathbb{R}^p$ is a random vector whose distribution depends on the query $q$ and the generative model $f$. In reality, we do not know the distribution of $g(f(q))$, hence we obtain iid replicates $g(f(q)_1),\dots,g(f(q)_r)$ from which we empirically estimate the response distribution and use it for subsequent inference. 
\newline
\newline
In order to conduct statistical inference on a set of black-box generative models, several works in the literature including \cite{helm2024tracking} suggests embedding the given set of generative models into a finite-dimensional Euclidean space. In particular, \cite{helm2024tracking} suggests performing multidimensional scaling upon the pairwise dissimilarities between the distributions of the sample responses of the generative models to some user-given queries, to obtain a vector representation of every generative model in the given set of generative models.  This method is termed \textit{Data Kernel Perspective Space embedding}. 
\newline
\newline
\cite{helm2024embedding} and \cite{acharyya2024consistent} investigate the asymptotic properties of the sample vector embeddings, and establish consistency results under appropriate conditions.  
However, there is a tremendous increase in cost with the increase in sample size of responses from the generative models, 
making the asymptotic regime unrealistic. Hence, it is of interest to obtain high probability concentration bounds for the sample vector embeddings, which provides guarantees for finite-sample scenarios. To be precise, it can tell us that in order to reach a desired level of accuracy with a target level of confidence, what the minimum size of the sample of responses should be.
\newline
\newline
In the next section, we describe the \textit{Data Kernel Perspective Space embedding} method. 
\newline
\section{Description of the Data Kernel Perspective Space embedding method} 
\label{Sec:Decsription_of_DKPS}
Our setting involves a set of generative models $\lbrace f_1,\dots,f_n \rbrace$, each of which is a random map from a common query space $\mathcal{Q}$ to a common response space $\mathcal{X}$. Each generative model responds to a set of user-given queries, $\lbrace q_1,\dots,q_m \rbrace \subset \mathcal{Q}$. There is an embedding function $g$ which maps every response to a vector in $\mathbb{R}^p$. The distribution of the (vectorized) response of $f_i$ to $q_j$ is $F_{ij}$, that is, $g(f_i(q_j)) \sim F_{ij}$ for all $i,j$. 
\newline
\newline
In reality, $F_{ij}$ is supported on a bounded subset of $\mathbb{R}^p$, which implies that $F_{ij}$ has finite moments of every order. 
At the population-level, the generative model $f_i$ is represented by the matrix $\boldsymbol{\mu}_i=\left[
\mathbb{E}[g(f_i(q_1))] \big|
\mathbb{E}[g(f_i(q_2))] \big|
\dots \big|
\mathbb{E}[g(f_i(q_m))]
\right]^T \in \mathbb{R}^{m \times p}$ for all $i$, in the context of the given queries. The population-level pairwise  dissimilarities between the generative models are given by $\boldsymbol{\Delta}_{i i'}=\frac{1}{\sqrt{m}} \left\lVert 
\boldsymbol{\mu}_i-
\boldsymbol{\mu}_{i'}
\right\rVert_F$. Classical Multidimensional Scaling of the pairwise dissimilarities $\boldsymbol{\Delta}_{i i'}$ into $\mathbb{R}^d$ yields vectors $\boldsymbol{\psi}_1,\dots,\boldsymbol{\psi}_n \in \mathbb{R}^d$. 
\newline
\newline
In reality, we cannot compute the vectors $\boldsymbol{\psi}_i$ because the response distributions $F_{ij}$ are unknown. Hence, we obtain iid replicates of responses for each model to every query, that is, for every $i,j$, we obtain 
\begin{equation*}
    g(f_i(q_j)_1),\dots,g(f_i(q_j)_r) \sim^{iid} F_{ij},
\end{equation*}
where $f_i(q_j)_k$ denotes the $k$-th replicate of the response of $f_i$ to $q_j$. 
We estimate $\boldsymbol{\mu}_i$ with its sample counterpart $\bar{\mathbf{X}}_i=
\left[
\frac{1}{r}
\sum_{k=1}^r g(f_i(q_1)_k)
\big|
\frac{1}{r}
\sum_{k=1}^r 
g(f_i(q_2)_k)
\big|
\dots|
\frac{1}{r}
\sum_{k=1}^r 
g(f_i(q_m)_k)
\right]^T$. We perform Classical Multidimensional Scaling upon the sample dissimilarity matrix $\mathbf{D}$ given by 
\begin{equation*}
    \mathbf{D}=
    \left(
\frac{1}{\sqrt{m}}
\left\lVert 
\bar{\mathbf{X}}_i-
\bar{\mathbf{X}}_{i'}
\right\rVert_F
\right)_{i,i'=1}^n,
\end{equation*}
to obtain the sample vector embeddings $\lbrace \hat{\boldsymbol{\psi}}_1,\dots,\hat{\boldsymbol{\psi}}_n \rbrace$. 
\newline
We describe the algorithm for the abovementioned procedure, referred to as \textit{Data Kernel Perspective Space embedding}, in  \Cref{Algo:DKPS} below.
\begin{algorithm}[H]
\caption{DKPSembed($
\lbrace 
f_i
\rbrace_{i=1}^n;
\lbrace
q_j
\rbrace_{j=1}^m;r;g;d
$)}. 
\label{Algo:DKPS}
\begin{algorithmic}[1]
\State Generate $r$ independent and identically distributed replicates of responses from every model $f_i$ to every query $q_j$, that is, obtain
\begin{equation*}
    g(f_i(q_j)_1),\dots,g(f_i(q_j)_r) \sim^{iid} F_{ij}
    \text{ for all $i,j$}. 
\end{equation*}
\State For all $i \in [n]$, compute
$
\bar{\mathbf{X}}_i=
\bigg[
\frac{1}{r} \sum_{k=1}^r 
g(f_i(q_1)_k)
\bigg|
\frac{1}{r} \sum_{k=1}^r 
g(f_i(q_2)_k)
\bigg|
\dots 
\bigg| \frac{1}{r} \sum_{k=1}^r 
g(f_i(q_m)_k)
\bigg]^T \in \mathbb{R}^{m \times p}
$.
\State Compute the sample dissimilarity matrix,
$
\mathbf{D}=
\bigg(
\frac{1}{\sqrt{m}}
\left\lVert
\bar{\mathbf{X}}_i-
\bar{\mathbf{X}}_{i'}
\right\rVert_F
\bigg)_{i,i'=1}^n
$.
\State Obtain 
$\hat{\mathbf{B}}=-\frac{1}{2}
\mathbf{H}_n \mathbf{D}^{\circ 2} \mathbf{H}_n^T$,  compute $\hat{\mathbf{U}}^{(d)}=\left[
\mathbf{u}_1(\hat{\mathbf{B}}) \big|\dots \big|
\mathbf{u}_d(\hat{\mathbf{B}})
\right]$ and $\hat{\mathbf{S}}^{(d)}=\mathrm{diag}(\lambda_1(\hat{\mathbf{B}}),\dots,\lambda_d(\hat{\mathbf{B}}))$, and subsequently calculate $\hat{\boldsymbol{\psi}}=\hat{\mathbf{U}}^{(d)} \left( \hat{\mathbf{S}}^{(d)} \right)^{\frac{1}{2}} \in \mathbb{R}^{n \times d}$.   
\State \Return $\lbrace 
\hat{\boldsymbol{\psi}}_1,\dots \hat{\boldsymbol{\psi}}_n
\rbrace$ where $\hat{\boldsymbol{\psi}}_i$ is the $i$-th row of $\hat{\boldsymbol{\psi}}$.
\end{algorithmic}
\end{algorithm}
\begin{remark}
    We shall henceforth refer to the vector 
    $\boldsymbol{\psi}_i$
    as the true perspectives
    of the generative model $f_i$, and $\hat{\boldsymbol{\psi}}_i$ as the estimated perspective of $f_i$ for all $i$ (in the context of the set of queries $\lbrace q_j \rbrace_{j=1}^m$). 
\end{remark}
For sake of convenience, we shall henceforth use the notations $\mathbf{B}=-\frac{1}{2} \mathbf{H}_n
\boldsymbol{\Delta}^{\circ 2} \mathbf{H}_n^T
$, and $\hat{\mathbf{B}}=-\frac{1}{2} \mathbf{H}_n^T \mathbf{D}^{\circ 2} \mathbf{H}_n^T$.
Moreover, $\lambda_1,\lambda_2 ,\dots,\lambda_n$ denote the eigenvalues of $\mathbf{B}$ arranged in decreasing order of magnitude (that is, $|\lambda_1| \geq |\lambda_2| \geq \dots \geq |\lambda_n|$), and $\hat{\lambda}_1,\hat{\lambda}_2, \dots, \hat{\lambda}_n$
denote the eigenvalues of $\hat{\mathbf{B}}$ arranged in decreasing order of magnitude (that is, $|\hat{\lambda}_1| \geq |\hat{\lambda}_2| \geq \dots \geq |\hat{\lambda}_n|$).
\newline
\newline
In the following sections, we establish a high probability concentration bound on the estimation error of the perspectives, and  demonstrate its applicability in a subsequent inference task.  
\newline 
\section{Theoretical results}
\label{Sec:Theoretical_results}
We present our theoretical results in this section. 
Our key concentration inequality result holds for any finite collection of generative models, but since we want the estimation error to be smaller with high probability as the number of generative models ($n$) grows, we  show that our estimation error approaches zero in the asymptotic regime. Since we draw our inferences on the generative models from the sample responses, we expect that the number of sample responses to every query ($r$) must also grow sufficiently fast as the number of generative models increases, to ensure that the estimation error approaches zero with high probability. 
\newline
\newline
Recall that the estimated perspectives $\hat{\boldsymbol{\psi}}_i$ are obtained by Classical Multidimensional Scaling of the (doubly centered) sample dissimilarity matrix $\hat{\mathbf{B}}$ (while the true perspectives $\boldsymbol{\psi}_i$ are Classical Multidimensional Scaling outputs of $\mathbf{B}$). Thus, in order to establish a concentration bound on the error of estimation of $\boldsymbol{\psi}$ by $\hat{\boldsymbol{\psi}}$, it would be helpful to obtain a concentration bound on $\left\lVert \hat{\mathbf{B}}-\mathbf{B} \right\rVert$.
However, in order to establish a concentration bound on $\left\lVert
\hat{\mathbf{B}}-\mathbf{B}
\right\rVert$, we first need to make a distributional assumption on the responses. 
\begin{assumption}
\label{Asm:Response_distribution}
For all $i \in [n]$, define $\mathbf{x}_i=
\left[
(\bar{\mathbf{X}}_i)^T_{1 \cdot},
(\bar{\mathbf{X}}_i)^T_ {2 \cdot},\dots,
(\bar{\mathbf{X}}_i)^T_{m \cdot}
\right]^T \in \mathbb{R}^{mp}
$ to be the concatenation of the sample mean (vectorized) responses from $f_i$. Then, the following two conditions hold simultaneously:
\begin{enumerate}
    \item For all $n,m,r \in \mathbb{N}$, for all $i \in [n]$, any two entries of $\mathbf{w}_i=(\mathrm{cov}(\mathbf{x}_i))^{-\frac{1}{2}}(\mathbf{x}_i-\mathbb{E}(\mathbf{x}_i))$ are uncorrelated.
\item There exists $\omega>0$ such that for all $n,m,r \in \mathbb{N}$, for all $i \in [n]$, the density of the random vector $\mathbf{w}_i$ is proportional to $e^{-U(\mathbf{w})}$ where $\nabla^2 U(\mathbf{w}) \geq 
\frac{1}{\omega^2}
\mathbf{I}_{mp}$.
\item There exists an $L>0$ such that for all $n,m,r \in \mathbb{N}$, for all $i \in [n]$, $\left\lVert 
\mathbf{x}_i
\right\rVert \leq L$ almost surely. 
\end{enumerate}
\end{assumption}
Observe that  \Cref{Asm:Response_distribution} states that the standardized version of the concatenated mean responses are light-tailed, and have uncorrelated entries. If every response distribution is truncated multivariate normal, then this assumption holds \citep{amini2021concentration}. Moreover, if a large number of iid replicates of responses are sampled from every response distribution, then every $\mathbf{x}_i$ has a distribution that is close to multivariate normal by Central Limit Theorem, in addition to the fact that any vectorized response from a large language model is always uniformly bounded due to finiteness of number of tokens used to construct the response, thus making our assumption realistic.  
Next, we state our concentration bound on the distance between the doubly centered dissimilarity matrices. 
\begin{theorem}
\label{Thm:Noise_CMDS_dissimilarity_bound}
    In our setting,
    suppose \textit{Assumption \ref{Asm:Response_distribution}} holds.
    Define
    $\boldsymbol{\Sigma}_{ij}$ to be the covariance matrix associated with the probability distribution $F_{ij}$ of responses from $f_i$ to $q_j$, and subsequently define $\gamma_{ij}=\mathrm{trace}(\boldsymbol{\Sigma}_{ij})$.
    Assume that for all $i,j$,
    $\gamma_{ij} \leq \Gamma$ for some $\Gamma>0$.
    Then, there exist constants $C,c>0$ such that with probability at least $(1-e^{-n c})$,
    \begin{equation*}
    \left\lVert 
\hat{\mathbf{B}}-
\mathbf{B}
    \right\rVert
    \leq \frac{(2 L \omega C \sqrt{\Gamma}) n}{\sqrt{m r}}+\frac{\Gamma}{r}.
    \end{equation*}
\end{theorem}
Since $\gamma_{ij}$ is the trace of the dispersion matrix of the distribution of responses of $f_i$ to $q_j$, it 
denotes a measure for variability of the response distribution of $f_i$ to $q_j$. That is, a large value of $\gamma_{ij}$ is associated with a large variation amongst the responses of $f_i$ to $q_j$. 
\textit{Theorem \ref{Thm:Noise_CMDS_dissimilarity_bound}} establishes a bound on the spectral norm $\left\lVert \hat{\mathbf{B}}-
\mathbf{B} \right\rVert$, under the condition that the variability $\gamma_{ij}$ of the response distributions $F_{ij}$ are uniformly bounded by a constant, and the sample size $r$ grows faster than $n^2$. 
Note that since all $\gamma_{ij}$ are uniformly bounded by the constant $\Gamma$, the number of queries, $m$, does not matter. 
\newline
\newline
The spectral norm bound established in  \Cref{Thm:Noise_CMDS_dissimilarity_bound}, can be used to establish concentration bound on the estimation error $\left\lVert 
\hat{\boldsymbol{\psi}} \mathbf{W}_*-\boldsymbol{\psi}
\right\rVert$, using Weyl's Inequality, which puts a bound on the eigenvalue perturbations, and Davis Kahan Theorem, which puts a bound on the eigenvector perturbations. We make extensive use of the results in \cite{agterberg2022joint}, where $\hat{\boldsymbol{\psi}} \mathbf{W}_*-\boldsymbol{\psi}$ is decomposed into a sum of matrices. We establish a bound on each summand matrix, and thus establish our key concentration result.  
\newline
\newline
Prior to stating our key concentration result, we discuss the assumptions under which our result holds.  
\begin{assumption}
\label{Asm:constant_rank}
For all $n$ sufficiently large, $\mathrm{rank}(\mathbf{B})=d$ where $d$
is a constant not changing with $n$. 
\end{assumption}
Note that  \Cref{Asm:constant_rank} states that the
true perspectives $\boldsymbol{\psi}_i$ of the generative models reside in a $d$-dimensional Euclidean space. 
 \Cref{Asm:constant_rank} is based on our observation that in simulation and synthetic data analysis, we find that every scree plot of the (doubly centered) population dissimilarity matrices has an elbow at approximately the same value. It justifies our choosing the same embedding dimension as the number of generative models under consideration grows. 
\newline
\newline
While  \Cref{Asm:constant_rank} is not a strict necessity for deriving concentration bounds on the sample embeddings, it does facilitate convenient analysis and interpretation. Consider, for instance, a regime where the $\mathrm{rank}(\mathrm{\mathbf{B}})$ grows with $n$, thus violating  \Cref{Asm:constant_rank}. In such case, the perspective of a specific generative model has a growing number of components as more generative models are brought under consideration. This makes comparison across different sub-regimes difficult. For instance, if one wants to study the change in perspective of a specific generative model as $n$ grows, it is inconvenient to deal with vectors of growing dimensions. Nonetheless, we recognize that  \Cref{Asm:constant_rank} needs to be relaxed for taking into account more generalized scenarios, and we leave that to future work. 
\begin{proposition}
\textit{Assumption \ref{Asm:constant_rank}}  holds when every generative model $f_i$ is associated with a vector $\boldsymbol{\phi}_i$ on a  $d$-dimensional compact Riemannian manifold $\mathcal{M}$ in a high-dimensional ambient space $\mathbb{R}^q$, such that the pairwise geodesic distances 
$d_{\mathcal{M}}(\boldsymbol{\phi}_i,\boldsymbol{\phi}_{i'})$
equates to
the pairwise population-level dissimilarities between the mean (vector-embedded) responses of $f_i$ and $f_{i'}$, given by $\boldsymbol{\Delta}_{i i'}$. 
\end{proposition}
Next, we establish our main result stating a high probability concentration bound on the error of estimating $\boldsymbol{\psi}$ with $\hat{\boldsymbol{\psi}}$ (upto an orthogonal transformation). 
\begin{theorem}
\label{Th:Noisy_CMDS_concentration_bound}
In our setting, 
assume $\frac{(2 L \omega C \sqrt{\Gamma}) n}{\sqrt{m r}} +\frac{\Gamma}{r} < \frac{|\lambda_d|}{2} $, and
suppose
\textit{Assumption \ref{Asm:Response_distribution}} and \textit{Assumption \ref{Asm:constant_rank}} hold. Then, with probability at least
$(1-13 e^{-n c})$, there exists an $\mathbf{W}_* \in \mathcal{O}(d)$ such that
\begin{equation*}
\begin{aligned}
\left\lVert
\hat{\boldsymbol{\psi}} \mathbf{W}_*-
\boldsymbol{\psi}
\right\rVert \leq
\left\lbrace 
(2+\sqrt{2} + 5 \sqrt{2} d +4 d \sqrt{\kappa})
\right\rbrace
\frac{1}{|\lambda_d|^{\frac{1}{2}}}
\left(
\frac{(2 L \omega C \sqrt{\Gamma}) n}{\sqrt{m r}} + 
\frac{\Gamma}{r}
\right),
\end{aligned}
\end{equation*}
$\kappa=\frac{|\lambda_1|}{|\lambda_d|}$ being the condition number of $\mathbf{B}$.
\end{theorem}
Thus, in a setting where the response distributions have uniformly bounded variability (that is, $\max_{i \in [n],j \in [m]} \gamma_{ij}=O(1)$), the abovementioned  \Cref{Th:Noisy_CMDS_concentration_bound} yields a high probability concentration bound for the sample vector embeddings of a finite number of generative models. 
Note that the coefficient in  \Cref{Th:Noisy_CMDS_concentration_bound} depends on $n$. Assuming that the coefficient is bounded above uniformly while $n$ varies, observe that the estimation error approaches zero if $n=o(\sqrt{m r})$. If we further restrict ourselves to a regime where the number of queries, $m$, remains constant, then $r=\omega(n^2)$ ensures consistent estimation of the perspectives $\boldsymbol{\psi}_i$.  
\newline
\newline
Note that the coefficients of the polynomial bound in  \Cref{Th:Noisy_CMDS_concentration_bound} are expected to change with $n$. While  \Cref{Asm:constant_rank} ensures that $d$ remains constant, it can be seen from  \Cref{Th:Noisy_CMDS_concentration_bound} that the eigenvalues $\lambda_1$ and $\lambda_d$ control the error of estimating $\boldsymbol{\psi}$ with $\hat{\boldsymbol{\psi}}$. In order to ensure that the estimation error approaches zero, the growth of the eigenvalues need to be controlled. 
Below, we state a result discussing a sufficient condition for favorable growth of the eigenvalues of $\mathbf{B}$, so that consistency of the $\hat{\boldsymbol{\psi}}$ is ensured. 
\begin{proposition}
    \label{Prop:suff_for_growing_eigenvalues}
    In our setting, suppose the following conditions hold simultaneously:
    \begin{enumerate}
        \item In the collection of vectorized population  mean response  matrices
        $\left\lbrace
        \tilde{\boldsymbol{\mu}}_i
\right\rbrace_{i=1}^n$, where 
$\tilde{\boldsymbol{\mu}}_i=
\left[
(\boldsymbol{\mu}_i)_{1 \cdot}^T,
\dots 
(\boldsymbol{\mu}_i)_{m \cdot}^T
\right]^T
$,
there are at most $K$ distinct members, denoted by $\tilde{\boldsymbol{\mu}}^{(1)},\tilde{\boldsymbol{\mu}}^{(2)},\dots,
\tilde{\boldsymbol{\mu}}^{(K)}$, where $K \geq d$. 
\item Denoting by $n_i$ the number of generative models with population mean response matrix $\boldsymbol{\mu}_i$, $n_i=O(n)$ for all $i \in [K]$. 
    \end{enumerate}
    Then, there exists a constant $C^{(d)}>0$ such that $\lambda_d(\mathbf{B}) \geq C^{(d)} n$ for all sufficiently large $n$.
\end{proposition}
Observe that  \Cref{Prop:suff_for_growing_eigenvalues} states a sufficient condition for the signal contained in every component of the true perspectives $\boldsymbol{\psi}_i$ to grow with $n$, which ensures the consistency of $\hat{\boldsymbol{\psi}}$ when $n=o(\sqrt{m r})$.  
From an intuitive standpoint, 
estimation of the perspectives become more difficult as the number of generative models increases. Hence, it is expected that the amount of signal contained in every component of the true perspectives $\boldsymbol{\psi}_i$ should increase with $n$, in order to facilitate consistent estimation of the true perspectives. Consider, for instance, a regime where $\lambda_d$ is arbitrarily small. That would have made the coefficients in the polynomial bound on the estimation error in  \Cref{Th:Noisy_CMDS_concentration_bound} blow up infinitely often. 
\newline
\newline
Note that the conditions mentioned in  \Cref{Prop:suff_for_growing_eigenvalues} hold 
 when there are only a
finite number of distinct communities in terms of the population mean response matrices, and the membership for every community is proportional to the total number of generative models. Thus,
 when there are only a finite number of pretrained generative models, and from each pretrained model a growing number of generative models are finetuned, we can expect the estimated perspectives $\hat{\boldsymbol{\psi}}$ to be consistent.  
\newline
\newline
\newline
After presenting our theoretical results in the current section, we back them up with our results from numerical experiments in the next section. 
\section{Numerical Experiments}
We present our numerical results in this section. At first, we present our simulations in  \Cref{Subsec:Simulations}. In our simulations, we simulate the responses from a large language model with high dimensional random vectors, and demonstrate that our concentration bounds exhibit high empirical coverage. Then, in \Cref{Subsec:Real_Data_Analysis}, we generate responses from an actual large language model to demonstrate that the concentration bound is satisfied on all the instances of sampling.  
\label{Sec:Numerical_Experiments}
\subsection{Simulations}
\label{Subsec:Simulations}
Here, we simulate a large language model which outputs random vectors of dimension $p=75$, with a random number generator.
Our goal is to show that under appropriate conditions, as the number of generative models, $n$, increases, the estimation error $\arg \min_{\mathbf{W}_* \in \mathcal{O}(d)} \left\lVert
\hat{\boldsymbol{\psi}} \mathbf{W}_* -
\boldsymbol{\psi}
\right\rVert$ decreases, and so does the upper bound on the estimation error as stated in  \Cref{Thm:Noise_CMDS_dissimilarity_bound}.
\newline
\newline
We vary the number of generative models, $n$ in $\lbrace 5, 10,15,20,25 \rbrace$,  and take the number of iid replicates as $r=\lfloor n^{2.25} \rfloor$. We always keep the number of queries $m=2$ constant. 
For every $n \in \lbrace 5,10,15,20,25 \rbrace$, we do the following. 
We first generate $K=4$ matrices $\left\lbrace \tilde{\boldsymbol{\mu}}^{*(l)}
\right\rbrace_{l=1}^4$ whose each row is within $[-2,2]^p$. 
Then, for every $l \in [K]$, we create a matrix $\boldsymbol{\mu}^{(l)}$ whose $j$-th row $(\tilde{\boldsymbol{\mu}}^{(l)})_{j \cdot}$ is obtained by taking the empirical mean over $R=\lfloor n^{3.25} \rfloor$ iid random vectors, generated from a normal distribution with mean $(\tilde{\boldsymbol{\mu}}^{*(l)})_{j \cdot}$ and then truncated within $[-2,2]^p$. These matrices $\tilde{\boldsymbol{\mu}}^{(l)}$ represent the distinct members of the collection of all matrices of the population mean responses of the $n$ generative models.
\newline
\newline
Then, we repeat the following procedure for every $n \in \left\lbrace 5,10,15,20,25 \right\rbrace$. We assign the finite collection $\left\lbrace
\tilde{\boldsymbol{\mu}}^{(l)}
\right\rbrace_{l=1}^K$
to $n$ different generative models as their matrices of population mean responses.
For that purpose, for all $i \in [n]$, we define $\boldsymbol{\mu}_i=\tilde{\boldsymbol{\mu}}^{(i \text{ mod } K)}$. We subsequently obtain the population dissimilarity matrix
\begin{equation*}
    \boldsymbol{\Delta}=
    \left(
\frac{1}{\sqrt{m}}
\left\lVert 
\boldsymbol{\mu}_i-
\boldsymbol{\mu}_{i'}
\right\rVert_F
    \right)_{i,i'=1}^n
\end{equation*}
and obtain the matrix of the population-level embeddings, $\boldsymbol{\psi} \in \mathbb{R}^{n \times d}$ by performing Classical Multidimensional Scaling on $\boldsymbol{\Delta}$.
We find that for all values of $n$, the rank of the doubly centered population dissimilarity matrix
$\mathbf{B}=-\frac{1}{2} \mathbf{H}_n \boldsymbol{\Delta}^{\circ 2} \mathbf{H}_n$
is $3$, hence we choose $d=3$ as the embedding dimension.
\newline
\newline
Next, we describe how we obtain the matrix of the sample embeddings $\hat{\boldsymbol{\psi}}$. For each of $100$ Monte Carlo samples, we generate matrices $\left\lbrace \bar{\mathbf{X}}_i
\right\rbrace_{i=1}^n$ , where the $j$-th row of $\bar{\mathbf{X}}_i$ is obtained by the empirical mean of $r=\lfloor n^{2.25} \rfloor$ iid random vectors, generated from a Normal distribution with mean $(\tilde{\boldsymbol{\mu}}^{*(i \text{ mod } K)})_{j \cdot}$
and covariance matrix 
$\boldsymbol{\Sigma}^{(\text{resp})}$, and then truncated within $[-2,2]^p$, which ensures that $\mathbb{E}(\bar{\mathbf{X}}_i)
\approx
\boldsymbol{\mu}_i$.
We obtain the covariance matrix
$\boldsymbol{\Sigma}^{(\text{resp})}=0.005 \times
\mathbf{L}^{(\text{resp})} (\mathbf{L}^{(\text{resp})})^T$,
where $\mathbf{L}^{(\text{resp})}$ is a lower triangular matrix whose non-zero entries
are independently and identically distributed as $\mathrm{Unif}(0.25,0.30)$.
We compute the sample dissimilarity matrix $\mathbf{D}$ and subsequently obtain the matrix of the sample-level embeddings, $\hat{\boldsymbol{\psi}}$ by performing Classical Multidimensional Scaling on $\mathbf{D}$. After computing $\hat{\boldsymbol{\psi}}$ on each of the $100$ Monte Carlo samples, we compute the average of the quantity $\min_{\mathbf{W_* \in \mathcal{O}(3)}}
\left\lVert 
\hat{\boldsymbol{\psi}} \mathbf{W}_*-
\boldsymbol{\psi}
\right\rVert
$ over all the $100$ Monte Carlo samples, and plot it against $n$. We find that this estimation error decreases steadily with increase in $n$. 
\newline
\newline
We also find the value of the upper bound on concentration (right hand side of the equation in \Cref{Th:Noisy_CMDS_concentration_bound}) for each $n$. Since every response is within $[-2,2]^p$, we take the upper bound $L=2p$. Additionally, we take $\omega=1$ and $C=0.0005$. We take $\Gamma=\mathrm{trace}(\boldsymbol{\Sigma}^{(\text{resp})})$.
We compare the upper bound against the estimation error for each $n$ in \Cref{Tab:simulation}.
We find that the upper bound steadily decreases with increase in $n$.
\newline
\begin{figure}[htbp]
    \centering
    \begin{subfigure}
        \centering
    \includegraphics
        [scale=0.65]{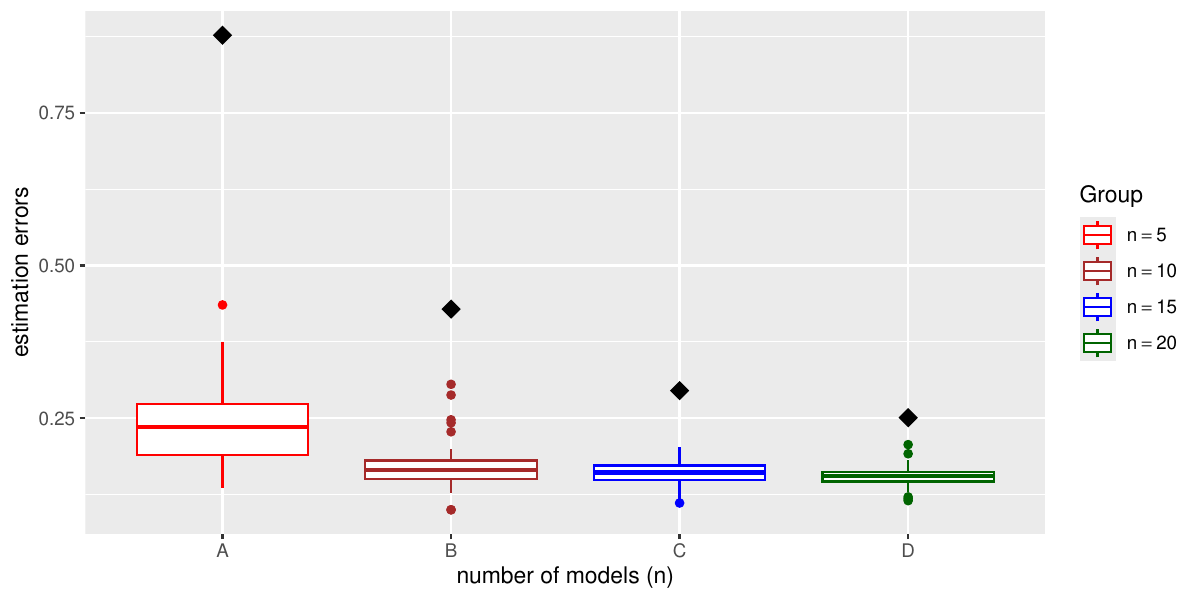}
    \caption*{Boxplot of estimation errors for perspectives of generative models when the number of distinct population mean response matrices is constant at $K=3$.}
    \end{subfigure}
    \begin{subfigure}
        \centering
    \includegraphics
        [scale=0.65]{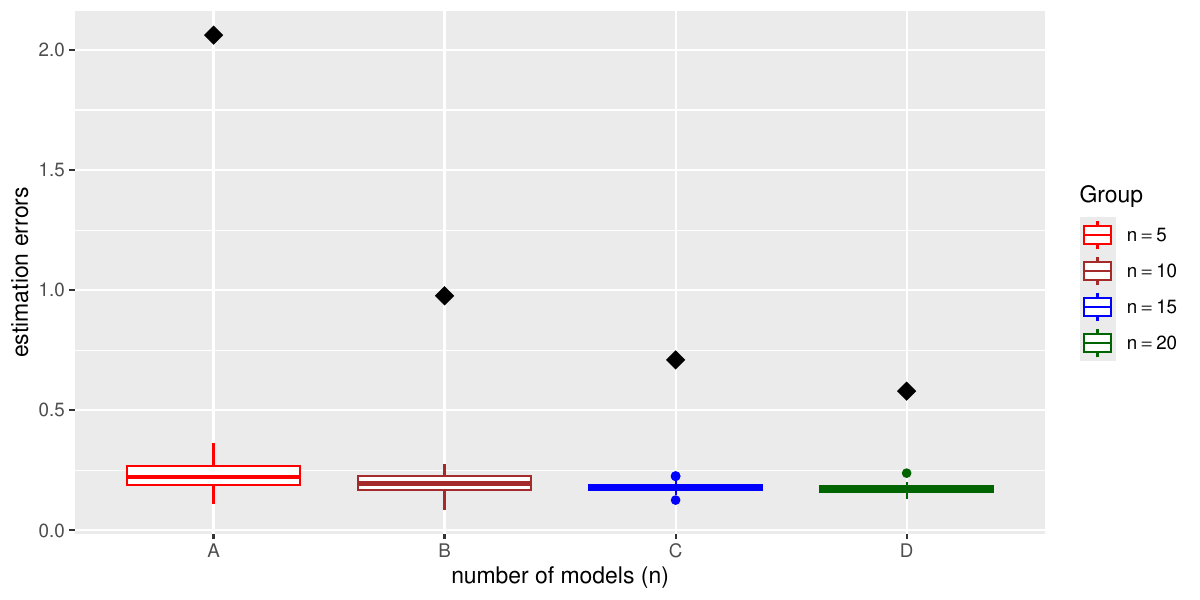}
    \caption*{Boxplot of estimation errors for perspectives of generative models when the number of distinct population mean response matrices is constant at $K=5$.}
    \end{subfigure}
    \caption{Boxplots showing the estimation errors at different values of $n$, the number of generative models. The upper panel corresponds to $K$, the number of distinct population mean response matrices, being equal to $3$, while the lower panel corresponds to $K=5$. No matter what the number of models is, the estimation errors are found to be always less than the suggested upper bound, marked by a black diamond. It can also be seen that as the number if generative models increase from left to right, the variation in the values of the estimation error decreases. }
\end{figure}

Then, we compute the values of $\arg \min_{\mathbf{W} \in \mathcal{O}(d)}
\left\lVert
\hat{\boldsymbol{\psi}} \mathbf{W}- 
\boldsymbol{\psi}
\right\rVert
$ on $100$ Monte Carlo samples, and compute on what proportion of them 
the bound established in  \Cref{Th:Noisy_CMDS_concentration_bound}
is satisfied. 
\newline
\newline
We find that our bound is satisfied on $100 \%$ of the Monte Carlo samples, for every $n$. The bound appears to be sharp on our choice of hyperparameters.
\newline
\newline
\begin{table}
\begin{tabular}{ccccc}
    \toprule
    n & m & $\text{average}\min_{\mathbf{W} \in \mathcal{O}(d)}\|\hat{\boldsymbol{\psi}} \mathbf{W}-\boldsymbol{\psi}\|$ & \text{upper bound} & \text{Empirical Coverage}  \\
    \midrule
    5 & 2 & 0.235 & 1.590 & 100\% \\
    7 & 2 & 0.201 & 1.116 & 100\% \\
    10 & 2 & 0.190 & 0.774 & 100\% \\
    12 & 2 & 0.175 & 0.594 & 100\% \\
    15 & 2 & 0.162 & 0.528 & 100\% \\
    17 & 2 & 0.159 & 0.493 & 100\% \\
    20 & 2 & 0.153 & 0.419 & 100\% \\
    22 & 2 & 0.155 & 0.410 & 100\% \\
    25 & 2 & 0.149 & 0.376 & 100\% \\
    \bottomrule
\end{tabular}
\caption{Analysis of vector embeddings obtained by simulating responses from LLMs with high dimensional vectors,
based on $100$ Monte Carlo samples for each value of $n$.
The number of queries remain constant at $2$. The third column represents the average estimation error over all $100$ Monte Carlo samples, which is found to be sufficiently smaller than the suggested upper bound, satisfied on all the $100$ Monte Carlo samples.}
\label{Tab:simulation}
\end{table}
\newline
\newline

\subsection{Real Data Analysis}
\label{Subsec:Real_Data_Analysis}
We use \texttt{Google-gemma-2-2b-it} to run the experiments. We set $m=2$ where 
the queries are $q_1=\text{``How do outliers impact statistical results?''}$ and
\newline
$q_2=\text{``What role does education play in social mobility?''}$. 
\newline
We use \texttt{nomic-ai/nomic-embed-text-v2-moe} to transform the responses to vectors in $\mathbb{R}^{768}$.
We vary $n$, the number of LLMs, in the range $\lbrace 4,8,12,\dots, 40 \rbrace$ and use $r=\lfloor n^{2.75} \rfloor$ to be the number of replicates of responses sampled for the practitioner to compute $\hat{\boldsymbol{\psi}}$. 
\newline
\newline
For every $n \in \left\lbrace 4,8,12,16,\dots, 40 \right\rbrace$, we perform the following procedure. We first sample $R=5623$ replicates of responses (denoted by $\mathbf{x}_{ij1},\dots,\mathbf{x}_{ijR}$) for every LLM $f_i$ to every query $q_j$, which we use to compute the matrices $\boldsymbol{\mu}_i \in \mathbb{R}^{2 \times 768}$. We make 
sure that there are only four distinct members in the collection of population mean response matrices $\left\lbrace
\boldsymbol{\mu}_i
\right\rbrace_{i=1}^n$.
Subsequently, we compute the 
population dissimilarity matrix
$\boldsymbol{\Delta}$ and calculate 
$\boldsymbol{\psi}^* \in \mathrm{CMDS}(\boldsymbol{\Delta},d)$ where $d$ is the rank of the doubly centered dissimilarity matrix $\mathbf{B}=-\frac{1}{2}\mathbf{H}_n \boldsymbol{\Delta}^{\circ 2} \mathbf{H}_n$. It is found that $d=2$ for every $n$ in our case. 
Then,  we sample $r=\lfloor n^{2.75} \rfloor$ replicates of the responses with replacement from the original pool of $R$ generated responses ($\lbrace \mathbf{x}_{ij1},\dots,\mathbf{x}_{ijR} \rbrace$), for every LLM $f_i$ to every query $q_j$. We use the bootstrapped sample responses to compute the matrices $\bar{\mathbf{X}}_i \in \mathbb{R}^{2 \times 768}$, compute the sample dissimilarity matrix $\mathbf{D}$
and subsequently compute the sample embedding matrix $\hat{\boldsymbol{\psi}} \in \mathbb{R}^{n \times d}$. For each $n$, we compute the quantity $\arg \min_{\mathbf{W} \in \mathcal{O}(2)}\left\lVert 
\hat{\boldsymbol{\psi}} \mathbf{W}-
\boldsymbol{\psi}^*
\right\rVert$, and check if it is below   the suggested upper bound, which is computed from the Right Hand Side of \Cref{Th:Noisy_CMDS_concentration_bound}.
\newline
\newline
We tabulate our findings in  \Cref{Tab:real_data}.
\newline
\newline
\begin{table}
\begin{tabular}{ccccc}
    \toprule
    n & m & $\min_{\mathbf{W} \in \mathcal{O}(d)}\|\hat{\boldsymbol{\psi}} \mathbf{W}-\boldsymbol{\psi}^* \| $ & \text{upper bound} & \text{Bound satisfied}  \\
    \midrule
    8 & 2 & 0.0133 & 0.1054 & ``Yes'' \\
    12 & 2 & 0.0131 & 0.0424 & ``Yes'' \\
    16 & 2 & 0.0094 & 0.0255 & ``Yes'' \\
    20 & 2 & 0.0077 & 0.0184 & ``Yes'' \\
    24 & 2 & 0.0069 & 0.0146 & ``Yes'' \\
    28 & 2 & 0.0073 & 0.0122 & ``Yes'' \\
    32 & 2 & 0.0070 &
    0.0105 & ``Yes'' \\
    35 & 2 & 0.0064 & 0.0101 & ``Yes'' \\
    \bottomrule
\end{tabular}
\caption{Analysis of real data from large language model \texttt{Google-gemma-2-2b-it}. The number of queries remain the same ($2$). For each model, $R=5623$ responses are initially sampled, which serves as a proxy for the population of responses.
For each value of $n$,
first we compute the perspective matrix $\boldsymbol{\psi}^*$ from the $R=5623$ response vectors, which serves as the proxy for the true perspective matrix $\boldsymbol{\psi}$.
Then, we further sample $r=\lfloor n^{2.75} \rfloor$ vectors from the $R=5623$ vectors, and we compute the estimation error of the sample perspective matrix $\hat{\boldsymbol{\psi}}$ for estimating $\boldsymbol{\psi}^*$. 
The third column represents the estimation error $\left\lVert 
\hat{\boldsymbol{\psi}}
\mathbf{W}_* -
\boldsymbol{\psi}^*
\right\rVert$, which can be seen to be sufficiently smaller than the upper bound on the fourth column. For every value of $n$, on all the Monte Carlo samples the upper bound is found to be satisfied.}
\label{Tab:real_data}
\end{table}
\newline
\newline
\newline
\section{Discussion}
\label{Sec:Discussion}
To facilitate statistical analysis and inference on a given set of black-box generative models, various works in literature propose embedding every generative model in the given class into a finite-dimensional Euclidean space, based on their responses to user-given queries. The
vector embeddings thus obtained can be used for further downstream tasks such as providing performance guarantees or identification of models with sensitive information.
\cite{helm2024tracking} proposes one such embedding method, known as the \textit{Data Kernel Perspective Space (DKPS) Embedding}, which 
obtains a response-based vector embedding for every member of a given set of generative models, by using iid responses from  generative models to every query.
\newline
\newline
In this paper, we obtain high probability concentration bounds for the DKPS vector embeddings. We show that if the number of iid responses from a generative model to a query grows sufficiently faster than the number of generative models in the given set, then we can bound the error for estimation of the population-level vector embeddings with a quantity that is a polynomial function of a positive power of the quantity $\left( \frac{n}{\sqrt{m r}} \right)$. Our results are derived under the condition that the distributions of the responses of the generative models have uniformly bounded variability (that is, $\max_{i \in [n],j \in [m]} \gamma_{ij}=O(1)$). This condition is based on the fact that in reality, the tokens (which are the building blocks of a response from a generative model) are sampled from a finite pool of tokens. Additionally, our results are based on the signal-to-noise ratio condition that 
$\frac{(2 L \omega C \sqrt{\Gamma}) n}{\sqrt{ m r}}+\frac{\Gamma}{r}< \frac{|\lambda_d|}{2}$. While this is sufficient for our results to hold, a relevant question to ask is whether this condition can be relaxed. In settings that are much simpler to ours, such as the ones discussed in \cite{cai2018rate}, \cite{zhang2022heteroskedastic} and \cite{chen2021spectral}, a condition similar to ours is shown to be necessary. Since our setting is more complicated, we conjecture that our condition is also necessary. 
\newline
\newline
This gives us the ability to decide what the sample size $r$ should be, for a particular inference problem, for reaching a desired level of accuracy. 
First, note that a spectral norm bound is also a uniform bound, because $\left\lVert \hat{\boldsymbol{\psi}} \mathbf{W}_*-\boldsymbol{\psi} \right\rVert_{2,\infty} \leq \left\lVert 
\hat{\boldsymbol{\psi}} \mathbf{W}_*-\boldsymbol{\psi}
\right\rVert$. 
Hence, using our key result \Cref{Th:Noisy_CMDS_concentration_bound}, we can estimate how large $r$ should be in order to ensure that all the estimated perspectives $\hat{\boldsymbol{\psi}}_i$ are within a desired proximity of their population counterparts $\boldsymbol{\psi}_i$(up to a rotation), with high probability. This essentially lets us decide the sample size $r$ in order to obtain a desired level of accuracy on inference tasks (which are invariant to orthogonal transformations) involving the estimated perspectives $\hat{\boldsymbol{\psi}}_i$. 
Take, for instance, the problem of testing,  whether two specific generative models have the same perspective,
in a collection of $n$ generative models.
That is, without loss of generality, we want to test $H_0:\boldsymbol{\psi}_1=\boldsymbol{\psi}_2$. Clearly, we shall use the test statistic $T_{n,m,r}=\left\lVert \hat{\boldsymbol{\psi}}_1-\hat{\boldsymbol{\psi}}_2 \right\rVert$. Then, using  \Cref{Th:Noisy_CMDS_concentration_bound}, we can ensure $\left\lVert \hat{\boldsymbol{\psi}} \mathbf{W}_*-\boldsymbol{\psi} \right\rVert_{2,\infty} \leq \kappa$ for any desired $\kappa>0$, with high probability, by choosing a sufficiently large $r$. This means, by choosing a sufficiently large $r$, we can conclude that $\boldsymbol{\psi}_1 \neq \boldsymbol{\psi}_2$ with high probability, when we observe $\left\lVert 
\hat{\boldsymbol{\psi}}_1-\hat{\boldsymbol{\psi}}_2
\right\rVert>2 \kappa$. 
\newline
\newline
We discuss the significance and the scopes for future extension of our work in this paragraph. Primarily, as shown in  \Cref{Tab:simulation} and  \Cref{Tab:real_data}, the bounds for the estimation error $\min_{\mathbf{W} \in \mathcal{O}(d)}\left\lVert 
\hat{\boldsymbol{\psi}} \mathbf{W}-
\boldsymbol{\psi}
\right\rVert$ are sharp. 
 Moreover, the bound depends on the quantity $\frac{n}{\sqrt{m r}}$, leading us to the conclusion that we can increase the accuracy for estimating $\boldsymbol{\psi}_i$ not just by increasing the number of replicates ($r$), but also by increasing the number of queries ($m$). In order to control the cost of eliciting responses from a large language model, practitioners often increase only one of $m$ or $r$ but not both. However, as $m$ increases while $r$ remains fixed, the sample mean response matrices $\bar{\mathbf{X}}_i$ become poorer estimators of their population counterparts $\boldsymbol{\mu}_i$ (since every row $(\boldsymbol{\mu}_i)_{j \cdot}$ is estimated by $(\bar{\mathbf{X}}_i)_{j \cdot}$, but the number of rows increases as $m$ increases). Thus, it will be an interesting investigation to find the optimal choice for this trade-off. We would also like to draw the attention of the reader to the fact that the bound in \Cref{Th:Noisy_CMDS_concentration_bound} is satisfied when $\lambda_d(\mathbf{B})=\Omega(n)$, which holds when the number of distinct members in the collection of population mean response matrices $\left\lbrace \boldsymbol{\mu}_i
\right\rbrace_{i=1}^n$ is constant (see  \Cref{Prop:suff_for_growing_eigenvalues}). A possible future direction of research could investigate how to relax this condition, that will allow an increasing number of distinct members in the collection $\left\lbrace \boldsymbol{\mu}_i \right\rbrace_{i=1}^n$. 
We have not come across work of similar nature in the literature, pertaining to obtaining concentration bounds on response-based generative model embeddings. Since we consider response-based embeddings of generative models, it enables us to deal with generative models in a realistic black-box setting. The concentration bounds offer us finite-sample guarantees, facilitating theoretical foundation for study in the non-asymptotic regime.
\newline
\newline
\section{Code and Data availability}
All the codes and the data are available at the github repository \href{}{https://github.com/Aranyak-Acharyya/LLM-DKPS-Concentration.git}.
\newpage
\bibliographystyle{plainnat}
\bibliography{ref}
\newpage
\section{Appendix}
\label{Sec:Appendix}
In this section, we provide detailed proofs of our theoretical results. All our theoretical results (theorems, corollaries and propositions) in \Cref{Sec:Theoretical_results} are based on certain lemmas, and results from other papers in the literature. In \Cref{Subsec:Proofs_of_theorems}, we write the proofs of all the presented theoretical results, including theorems, corollaries and propositions stated in  \Cref{Sec:Theoretical_results}. In \Cref{Subsec:Proofs_of_supporting_propositions}, we provide proofs of supporting propositions which are not presented in the main body. In \Cref{Subsec:Proofs_of_lemmas}, we provide the proofs of the lemmas which our theoretical results depend on. Finally, in  \Cref{Subsec:Results_from_others}, we state the important results already existing in the literature that are pertinent to the deduction of our theoretical results.
\newline
\subsection{Proofs of presented results}
\label{Subsec:Proofs_of_theorems}
\textbf{Proof of \Cref{Thm:Noise_CMDS_dissimilarity_bound}.} 
Define $\Tilde{\boldsymbol{\mu}}_i=
\frac{1}{\sqrt{m}}
\left(
(\boldsymbol{\mu}_i)_{1 \cdot}^T,
(\boldsymbol{\mu}_i)_{2 \cdot}^T,
\dots,
(\boldsymbol{\mu}_i)_{m \cdot}^T
\right)^T$ and 
$\mathbf{x}_i=
\frac{1}{\sqrt{m}}
\left(
(\bar{\mathbf{X}}_i)_{1 \cdot}^T,
(\bar{\mathbf{X}}_i)_{2 \cdot}^T,
\dots,
(\bar{\mathbf{X}}_i)_{m \cdot}^T
\right)^T
$, for all $i \in [n]$. 
Subsequently, we define 
$\Tilde{\boldsymbol{\mu}}=
[\Tilde{\boldsymbol{\mu}}_1|\dots|\tilde{\boldsymbol{\mu}}_m]^T \in \mathbb{R}^{n \times mp}
$ and $\mathbf{X}=[\mathbf{x}_1|\dots|\mathbf{x}_n]^T \in \mathbb{R}^{n \times mp}$. Note that $\boldsymbol{\Delta}_{i i'}=\frac{1}{\sqrt{m}}
\left\lVert 
\boldsymbol{\mu}_i-
\boldsymbol{\mu}_{i'}
\right\rVert_F =
\left\lVert 
\tilde{\boldsymbol{\mu}}_i-
\tilde{\boldsymbol{\mu}}_{i'}
\right\rVert
$. 
\newline
Recall that $\mathbf{B}=-\frac{1}{2} \mathbf{H}_n \boldsymbol{\Delta}^{\circ 2} \mathbf{H}_n$, hence $\mathbf{B}=\mathbf{H}_n \tilde{\boldsymbol{\mu}}
\tilde{\boldsymbol{\mu}}^T 
\mathbf{H}_n^T
$. 
\newline
Similarly, we can prove $\hat{\mathbf{B}}=\mathbf{H}_n \mathbf{X} \mathbf{X}^T \mathbf{H}_n^T$. 
\newline
Thus, 
\begin{equation*}
\begin{aligned}
    \left\lVert 
\hat{\mathbf{B}}-\mathbf{B}
    \right\rVert &=
    \left\lVert 
-\frac{1}{2} \mathbf{H}_n \boldsymbol{\Delta}^{\circ 2}
\mathbf{H}_n^T -
(-\frac{1}{2}) 
\mathbf{H}_n \mathbf{D}^{\circ 2}
\mathbf{H}_n^T
    \right\rVert \\
    &= 
    \left\lVert 
\mathbf{H}_n 
\tilde{\boldsymbol{\mu}}
\tilde{\boldsymbol{\mu}}^T 
\mathbf{H}_n^T -
\mathbf{H}_n^T 
\mathbf{X}
\mathbf{X}^T 
\mathbf{H}_n^T
    \right\rVert \\
    &\leq 
    \left\lVert 
\tilde{\boldsymbol{\mu}}
\tilde{\boldsymbol{\mu}}^T
-
\mathbf{X}
\mathbf{X}^T
    \right\rVert \\
    &\leq 
    \left\lVert 
\mathbf{X} \mathbf{X}^T-
\mathbb{E} (\mathbf{X} \mathbf{X}^T)
    \right\rVert +
    \left\lVert 
\mathbb{E} (\mathbf{X} \mathbf{X}^T) -\tilde{\boldsymbol{\mu}}
\tilde{\boldsymbol{\mu}}^T
    \right\rVert.
\end{aligned}
\end{equation*}
\newline
Note that since a generative model can only generate a finite number of responses, there exists an $L>0$ such that for all $\mathbf{x} \in g(\mathcal{X})$, 
$\left\lVert \mathbf{x} \right\rVert \leq L$. 
\newline
This means, for all $i \in [n]$, for all $j \in [m]$, 
$\left\lVert (\bar{\mathbf{X}}_i)_{j \cdot} \right\rVert \leq L$.
\newline
Hence, for all $i \in [n]$,
\begin{equation*}
\begin{aligned}
    \left\lVert 
\mathbf{x}_i
    \right\rVert^2
    = \frac{1}{m}
    \sum_{j=1}^m \left\lVert 
(\bar{\mathbf{X}}_i)_{j \cdot}
    \right\rVert^2
    \leq L^2
    \implies 
    \left\lVert 
\mathbf{x}_i
    \right\rVert
    \leq L.
\end{aligned}
\end{equation*}
For all $\mathbf{w}_1,\mathbf{w}_2,\mathbf{z}_1,\mathbf{z}_2 \in 
\lbrace 
\mathbf{x}_i: i \in [n]
\rbrace
$, 
\begin{equation*}
\begin{aligned}
    \left|
\mathbf{w}_1^T \mathbf{w}_2-
\mathbf{z}_1^T \mathbf{z}_2
    \right|
    &\leq 
    \left|
    (\mathbf{w}_1-\mathbf{z}_1)^T 
    \mathbf{w}_2 +
    \mathbf{z}_1^T (\mathbf{w}_2-\mathbf{z}_2)
    \right| \\
    &\leq 
    \left|
    (\mathbf{w}_1-\mathbf{z}_1)^T 
    \mathbf{w}_2
    \right| +
    \left|
\mathbf{z}_1^T (\mathbf{w}_2-\mathbf{z}_2)
    \right| \\
    &\leq 
    \left\lVert 
\mathbf{w}_1-\mathbf{z}_1
    \right\rVert
    \left\lVert 
\mathbf{w}_2
    \right\rVert +
    \left\lVert 
\mathbf{z}_1
    \right\rVert
    \left\lVert
\mathbf{w}_2-\mathbf{z}_2
    \right\rVert
    \\
    &\leq 
    L 
    \left(
\left\lVert
\mathbf{w}_1-\mathbf{z}_1
\right\rVert+
\left\lVert
\mathbf{w}_2-\mathbf{z}_2
\right\rVert
    \right).
\end{aligned}
\end{equation*}
\newline
Define $\mathbf{w}_i=(\mathrm{cov}(\mathbf{x}_i))^{-\frac{1}{2}}(\mathbf{x}_i-\tilde{\boldsymbol{\mu}}_i)$ for all $i \in [n]$, and assume that \textit{Assumption \ref{Asm:Response_distribution}} holds for some $\omega>0$. 
\newline
Then, by Theorem-1 of \cite{amini2021concentration}, for all $t>0$,
\begin{equation*}
\begin{aligned}
\mathbb{P}
\left[
    \left\lVert 
\mathbf{X} \mathbf{X}^T -
\mathbb{E}(\mathbf{X} \mathbf{X})^T
    \right\rVert \leq 
    2 L
    \omega
    \sigma_{\infty}
    (Cn + \sqrt{n} t)
\right] \geq 1- e^{-\frac{t^2}{C^2}},
\end{aligned}
\end{equation*}
where $\sigma_{\infty}^2=\max_{i \in [n]}  \left\lVert 
\mathrm{cov}(\mathbf{x}_i)
\right\rVert
$.
\newline
If every model $f_i$ responds to every query $q_j$ independently, then, 
\begin{equation*}
\begin{aligned}
    \tilde{\boldsymbol{\Sigma}}_i=
    \mathrm{cov}(\mathbf{x}_i)=
    \frac{1}{m}
    \begin{pmatrix}
      \frac{1}{r}  \boldsymbol{\Sigma}_{i1} & \mathbf{0}^{p \times p} &  \dots
        & \mathbf{0}^{p \times p} 
        \\
    \mathbf{0}^{p \times p} & \frac{1}{r} \boldsymbol{\Sigma}_{i2} & \dots & 
    \mathbf{0}^{p \times p} \\
    \dots \\
    \dots \\
    \mathbf{0}^{p \times p} & \mathbf{0}^{p \times p} & \dots & 
    \frac{1}{r} \boldsymbol{\Sigma}_{im}
    \end{pmatrix}
    \implies 
    \left\lVert 
\tilde{\boldsymbol{\Sigma}}_i
    \right\rVert=
    \frac{1}{mr}
    \left(
    \max_{j \in [m]}
    \left\lVert 
\boldsymbol{\Sigma}_{ij}
    \right\rVert
    \right).
\end{aligned}
\end{equation*}
\newline
Now, note that for all $i,j$, $\left\lVert \boldsymbol{\Sigma}_{ij} \right\rVert \leq \gamma_{ij}$. Hence,
\begin{equation*}
\begin{aligned}
    \sigma_{\infty}^2 &=
    \max_{i \in [n]}
    \left\lVert 
\tilde{\boldsymbol{\Sigma}}_i
    \right\rVert \\
    &=
    \max_{i \in [n]}
    \left\lbrace 
\frac{1}{mr} 
\left(
\max_{j \in [m]}
\left\lVert 
\boldsymbol{\Sigma}_{ij}
\right\rVert
\right)
    \right\rbrace \\
    &\leq 
    \frac{1}{mr}
    \left(
    \max_{i \in [n], j \in [m]}
    \gamma_{ij}
    \right) \\
    &\leq 
    \frac{\Gamma}{mr}.
\end{aligned}
\end{equation*}
Thus, $\sigma_{\infty} \leq \sqrt{\frac{\Gamma}{mr}}$. 
\newline
\newline
Taking $t=\sqrt{n}$ in \textit{Theorem 1} of \cite{amini2021concentration}, and replacing $C$ with $(C-1)$ on the left hand side, we obtain, 
\begin{equation*}
\begin{aligned}
    \mathbb{P}
    \left[
\left\lVert
\mathbf{X} \mathbf{X}^T-
\mathbb{E}(\mathbf{X} \mathbf{X}^T)
\right\rVert
\leq
(2L \omega C \sqrt{\Gamma})
\frac{n}{\sqrt{mr}}
    \right]
    \geq 1 - e^{-\frac{n}{(C-1)^2}}.
\end{aligned}
\end{equation*}
Now,
\begin{equation*}
\begin{aligned}
    \mathbb{E}(\mathbf{X} \mathbf{X}^T)-\tilde{\boldsymbol{\mu}}
    \tilde{\boldsymbol{\mu}}^T &=
    \mathbb{E}(\mathbf{X}-\tilde{\boldsymbol{\mu}}+\tilde{\boldsymbol{\mu}})
    (\mathbf{X}-\tilde{\boldsymbol{\mu}}+\tilde{\boldsymbol{\mu}})^T
    -\tilde{\boldsymbol{\mu}}
    \tilde{\boldsymbol{\mu}}^T \\
    &= \mathbb{E}(\mathbf{X}-\tilde{\boldsymbol{\mu}})
    (\mathbf{X}-\tilde{\boldsymbol{\mu}})^T \\
    &= \frac{1}{m}
    \begin{pmatrix}
        \frac{1}{r}
        \sum_{j=1}^m \gamma_{1j}
        & 0 \dots & 0 \\
        0 & \frac{1}{r} \sum_{j=1}^m \gamma_{2j}
        \dots & 0 \\
        \dots \\
        0 & 0 \dots &
        \frac{1}{r}
        \sum_{j=1}^m \gamma_{nj}
    \end{pmatrix}.
\end{aligned}
\end{equation*}
Hence, 
\begin{equation*}
\begin{aligned}
    \left\lVert 
\mathbb{E}(\mathbf{X} \mathbf{X}^T)-
\tilde{\boldsymbol{\mu}}
\tilde{\boldsymbol{\mu}}^T
    \right\rVert &=
    \frac{1}{mr}
    \left(
    \max_{i \in [n]} \sum_{j=1}^m \gamma_{ij}
    \right)
    \leq \frac{\Gamma}{r}.
\end{aligned}
\end{equation*}
\newline
Hence, with probability at least $(1-e^{-\frac{n}{(C-1)^2}})$,
\begin{equation*}
\begin{aligned}
&\left\lVert 
\mathbf{X} \mathbf{X}^T -
\mathbb{E}(\mathbf{X} \mathbf{X}^T)
\right\rVert +
\left\lVert
\mathbb{E}(\mathbf{X} \mathbf{X}^T)- \tilde{\boldsymbol{\mu}} \tilde{\boldsymbol{\mu}}^T
\right\rVert 
\leq  \frac{(2 L \omega C \sqrt{\Gamma}) n}{\sqrt{m r}}+ \frac{\Gamma}{r} \\
&\implies 
\left\lVert 
\hat{\mathbf{B}}-\mathbf{B}
\right\rVert
\leq 
\frac{ (2 L \omega C \sqrt{\Gamma}) n}{\sqrt{m r}} +
\frac{\Gamma}{r}.
\end{aligned}
\end{equation*}
Taking $c=\frac{1}{(C-1)^2}$, we get the desired result. 
\newline
\newline
\newline
\newline
\textbf{Proof of \Cref{Th:Noisy_CMDS_concentration_bound}.} From Theorem \Cref{Thm:ref_Jagt_thm}, using Triangle Inequality, 
\begin{equation*}
\begin{aligned}
\left\lVert
\hat{\boldsymbol{\psi}} \mathbf{W}_*^T -
\boldsymbol{\psi}
\right\rVert
\leq 
\frac{1}{\sqrt{|\lambda_d|}}
\left\lVert
\hat{\mathbf{B}}-\mathbf{B}
\right\rVert +
\sum_{k=1}^6 \left\lVert 
\mathbf{R}_k
\right\rVert.
\end{aligned}
\end{equation*}
We know that, 
\begin{equation*}
\begin{aligned}
     &\left\lVert
\mathbf{R}_1
    \right\rVert
    \leq 
    \frac{\sqrt{2}}{\sqrt{|\lambda_d|}}
    \left(
\frac{(2 L \omega C \sqrt{\Gamma}) n}{\sqrt{m r}} + \frac{\Gamma}{r}
    \right)
\text{ with probability at least $(1-2 e^{-\frac{n}{(C-1)^2}})$}
    \text{ [from Proposition \Cref{Prop:R1_bound}]} \\
&\left\lVert
\mathbf{R}_2
    \right\rVert
    \leq 
\frac{\sqrt{2} }{(1+\sqrt{2})}
\frac{d}{\sqrt{|\lambda_d|}}
\left(
\frac{(2 L \omega C \sqrt{\Gamma}) n}{\sqrt{m r}}+\frac{\Gamma}{r}
\right)
\text{ with probability at least $(1-2 e^{-\frac{n}{(C-1)^2}})$}
\text{ [from Proposition \Cref{Prop:R2_bound}]}
  \\
&\left\lVert
\mathbf{R}_3
\right\rVert
\leq \frac{4d}{|\lambda_d|^2}
\sqrt{|\lambda_1|}
\left(
\frac{(2 L \omega C \sqrt{\Gamma}) n}{\sqrt{m r}}+\frac{\Gamma}{r}
\right)^2
\text{ with probability at least $(1- e^{-\frac{n}{(C-1)^2}})$}
\text{ [from Proposition \Cref{Prop:R3_bound}]}
\\
&\left\lVert \mathbf{R}_4 \right\rVert
\leq 
\frac{1}{|\lambda_d|^{\frac{3}{2}}}
\left(
\frac{(2 L \omega C \sqrt{\Gamma}) n}{\sqrt{m r}}+\frac{\Gamma}{r}
\right)^2
\text{ with probability at least $(1-e^{-\frac{n}{(C-1)^2}})$}
\text{ [from Proposition \Cref{Prop:R4_bound}]}
\\
&\left\lVert 
\mathbf{R}_5
    \right\rVert 
    \leq   
    \frac{2d}{(1+\sqrt{2})|\lambda_d|^{\frac{3}{2}}}
    \left(
\frac{(2 L \omega C \sqrt{\Gamma}) n}{\sqrt{m r}}+\frac{\Gamma}{r}
    \right)^2
    \text{ with probability at least $(1-4 e^{-\frac{n}{(C-1)^2}})$}
\text{ [from Proposition \Cref{Prop:R5_bound}]} \\
&\left\lVert 
\mathbf{R}_6
    \right\rVert
    \leq 
    \frac{4 \sqrt{2}d }{|\lambda_d|^{\frac{5}{2}}}
    \left(
\frac{(2 L \omega C \sqrt{\Gamma}) n}{\sqrt{m r}}+\frac{\Gamma}{r}
    \right)^3
    \text{ with probability at least $(1- 2 e^{-\frac{n}{(C-1)^2}})$}
     \text{ [from Proposition \Cref{Prop:R6_bound}]}.
\end{aligned}
\end{equation*}
Also, recall that
\begin{equation*}
\begin{aligned}
    \left\lVert 
\hat{\mathbf{B}}-\mathbf{B}
    \right\rVert
    \leq \left(
\frac{(2 L \omega C \sqrt{\Gamma}) n}{\sqrt{m r}} + 
\frac{\Gamma}{r}
    \right)
    \text{ with probability at least $(1-e^{-\frac{n}{(C-1)^2}})$}.
\end{aligned}
\end{equation*}
Combining the above bounds, $\text{with probability at least }
 (1-13 e^{-\frac{n}{(C-1)^2}})$,
\begin{equation*}
\begin{aligned}
    \left\lVert 
\hat{\boldsymbol{\psi}} \mathbf{W}_*^T -\boldsymbol{\psi}
    \right\rVert
    &\leq 
    \left\lbrace 
    \frac{1+\sqrt{2}}{|\lambda_d|^{\frac{1}{2}}}+
\frac{ \sqrt{2} d}{(1+\sqrt{2})|\lambda_d|^{\frac{1}{2}}}
    \right\rbrace
    \left(
\frac{(2 L \omega C \sqrt{\Gamma}) n}{\sqrt{m r}}+\frac{\Gamma}{r}
    \right) \\
    &\quad \quad + 
    \left\lbrace
    \frac{4d}{|\lambda_d|^2}
\sqrt{|\lambda_1|}
    +
    \frac{1}{|\lambda_d|^{\frac{3}{2}}}+
    \frac{2d}{(1+\sqrt{2})|\lambda_d|^{\frac{3}{2}}}
    \right\rbrace
    \left(
\frac{(2 L \omega C \sqrt{\Gamma}) n}{\sqrt{m r}}+
\frac{\Gamma}{r}
    \right)^2 +  
\frac{4 \sqrt{2}d}{|\lambda_d|^{\frac{5}{2}}}
\left(
\frac{(2 L \omega C \sqrt{\Gamma}) n}{\sqrt{m r}}+\frac{\Gamma}{r}
\right)^3
     \\
    &=
    \left\lbrace
    (1+\sqrt{2})
|\lambda_d|^{\frac{1}{2}} 
    +
    \frac{\sqrt{2} d}{(1+\sqrt{2})}
    |\lambda_d|^{\frac{1}{2}}
    \right\rbrace 
    \left\lbrace 
\frac{1}{|\lambda_d|}
\left(
\frac{(2 L \omega C\sqrt{\Gamma})n}{\sqrt{m r}} +
\frac{\Gamma}{r}
\right)
    \right\rbrace \\
    &\quad + 
    \left(
    4d \sqrt{|\lambda_1|}+
|\lambda_d|^{\frac{1}{2}} +
\frac{2d |\lambda_d|^{\frac{1}{2}}}{1+\sqrt{2}}
    \right)
    \left\lbrace 
\frac{1}{|\lambda_d|}
\left(
\frac{ (2 L \omega C \sqrt{\Gamma}) n}{\sqrt{m r}} +
\frac{\Gamma}{r}
\right)
    \right\rbrace^2 +
    4 \sqrt{2} d 
    |\lambda_d|^{\frac{1}{2}}
    \left\lbrace 
\frac{1}{|\lambda_d|}
\left(
\frac{(2 L \omega C \sqrt{\Gamma})n }{\sqrt{m r}}
\right)
    \right\rbrace^3
    \\
    &=
    \left\lbrace
    (1+\sqrt{2})
|\lambda_d|^{\frac{1}{2}} 
    +
    \frac{\sqrt{2} d}{(1+\sqrt{2})}
    |\lambda_d|^{\frac{1}{2}}
    \right\rbrace 
    \left\lbrace 
\frac{1}{|\lambda_d|}
\left(
\frac{(2 L \omega C\sqrt{\Gamma})n}{\sqrt{m r}} +
\frac{\Gamma}{r}
\right)
    \right\rbrace \\
    &\quad + 
    \left(
    4d \sqrt{\kappa}|\lambda_d|^{\frac{1}{2}}+
|\lambda_d|^{\frac{1}{2}} +
\frac{2d |\lambda_d|^{\frac{1}{2}}}{1+\sqrt{2}}
    \right)
    \left\lbrace 
\frac{1}{|\lambda_d|}
\left(
\frac{ (2 L \omega C \sqrt{\Gamma}) n}{\sqrt{m r}} +
\frac{\Gamma}{r}
\right)
    \right\rbrace^2
    \\
    &\quad 
    +
    4 \sqrt{2} d 
    |\lambda_d|^{\frac{1}{2}}
    \left\lbrace 
\frac{1}{|\lambda_d|}
\left(
\frac{(2 L \omega C \sqrt{\Gamma})n }{\sqrt{m r}}
\right)
    \right\rbrace^3
\end{aligned}
\end{equation*}
where $\kappa=\frac{|\lambda_1|}{|\lambda_d|}$.
\newline
Observe that 
\begin{equation*}
\begin{aligned}
    &\frac{(2 L \omega C \sqrt{\Gamma}) n}{\sqrt{ m r}} + 
    \frac{\Gamma}{r}
    \leq \frac{|\lambda_d|}{2} \\
    &\implies 
    \left\lbrace 
\frac{1}{|\lambda_d|}
\left(
\frac{(2 L \omega C \sqrt{\Gamma}) n}{\sqrt m r} + \frac{\Gamma}{r}
\right)
    \right\rbrace
    \geq 
    \left\lbrace 
\frac{1}{|\lambda_d|}
\left(
\frac{(2 L \omega C \sqrt{\Gamma}) n}{\sqrt m r} + \frac{\Gamma}{r}
\right)
    \right\rbrace^2 
    \geq 
    \left\lbrace 
\frac{1}{|\lambda_d|}
\left(
\frac{(2 L \omega C \sqrt{\Gamma}) n}{\sqrt m r} + \frac{\Gamma}{r}
\right)
    \right\rbrace^3.
\end{aligned}
\end{equation*}
Hence, $\text{with probability at least}  (1-13 e^{-n c})$ (and replacing $\mathbf{W}_*^T$ with $\mathbf{W}_*$),
\begin{equation*}
\begin{aligned}
    \left\lVert 
\hat{\boldsymbol{\psi}} \mathbf{W}_* -
\boldsymbol{\psi}
    \right\rVert
    &\leq 
    \left\lbrace
    \left(
(1+\sqrt{2})+ \frac{\sqrt{2} d}{(1+\sqrt{2})} +
4 d \sqrt{\kappa}
+ 1 +
\frac{2 d}{1+\sqrt{2}} +
4 \sqrt{2} d
    \right)
    |\lambda_d|^{\frac{1}{2}}
\right\rbrace 
\left\lbrace 
\frac{1}{|\lambda_d|}
\left(
\frac{(2 L \omega C \sqrt{\Gamma}) n}{\sqrt{m r}} + 
\frac{\Gamma}{r}
\right)
\right\rbrace \\
&=
\left\lbrace 
(2+\sqrt{2})+ 
5 \sqrt{2} d + 
4 d \sqrt{\kappa} 
\right\rbrace
\left\lbrace 
\frac{1}{|\lambda_d|^{\frac{1}{2}}}
\left(
\frac{ (2 L \omega C \sqrt{\Gamma}) n}{\sqrt{m r}} + 
\frac{\Gamma}{r}
\right)
\right\rbrace.
\end{aligned}
\end{equation*}
\newline
\newline
\newline
\textbf{Proof of \Cref{Prop:suff_for_growing_eigenvalues}.} 
First, we define the  population mean response vector for $f_i$ as $\tilde{\boldsymbol{\mu}}_i=
\left[
(\boldsymbol{\mu}_i)_{1 \cdot}^T,
(\boldsymbol{\mu}_i)_{2 \cdot}^T
\dots,
(\boldsymbol{\mu}_i)_{m \cdot}^T
\right]^T
\in \mathbb{R}^{mp}
$ and we subsequently define
$\tilde{\boldsymbol{\mu}}=
\left[
\tilde{\boldsymbol{\mu}}_i|
\dots|
\tilde{\boldsymbol{\mu}}_n
\right]^T \in \mathbb{R}^{n \times mp}
$.
\newline
Suppose, there are only $K$ distinct members in the collection of population mean responses $\left\lbrace
\tilde{\boldsymbol{\mu}}_1,
\tilde{\boldsymbol{\mu}}_2,
\dots,
\tilde{\boldsymbol{\mu}}_n
\right\rbrace$, denote them by 
$\tilde{\boldsymbol{\mu}}^{(1)},\dots,\tilde{\boldsymbol{\mu}}^{(K)}$. Then, $\tilde{\boldsymbol{\mu}}=\mathbf{Y} \boldsymbol{\mu}^*$, where $\boldsymbol{\mu}^*=[\tilde{\boldsymbol{\mu}}^{(1)}|
\dots
|
\tilde{\boldsymbol{\mu}}^{(K)}
]^T \in \mathbb{R}^{K \times mp}$, and $\mathbf{Y} \in \mathbb{R}^{n \times K}$ is defined as 
\begin{equation*}
    (\mathbf{Y})_{ij}=
    \mathbb{I}
    \left(    \tilde{\boldsymbol{\mu}}_i=\tilde{\boldsymbol{\mu}}^{(j)}
    \right)
\end{equation*}
for all $i \in [n], j \in [K]$. 
Note that 
\begin{equation*}
\begin{aligned}
    \lambda_d(\mathbf{B})
    &= \lambda_d(\mathbf{H}_n \tilde{\boldsymbol{\mu}} \tilde{\boldsymbol{\mu}}^T \mathbf{H}_n^T) \\ 
    &=
    \left(
    \sigma_d(\mathbf{H}_n \tilde{\boldsymbol{\mu}})
    \right)^2 \geq 
    \left( 
\sigma_d(\mathbf{H}_n) 
\sigma_d(\tilde{\boldsymbol{\mu}})
    \right)^2 \\ 
    &=
    \lambda_d(\mathbf{H}_n^T \mathbf{H}_n) \lambda_d(\tilde{\boldsymbol{\mu}} \tilde{\boldsymbol{\mu}}^T) \\
    &\geq  \lambda_d(\tilde{\boldsymbol{\mu}} \tilde{\boldsymbol{\mu}}^T) \\
    &= 
    \lambda_d
    (\mathbf{Y} \boldsymbol{\mu}^*
    \boldsymbol{\mu}^{*^T} \mathbf{Y}^T)
    \\
    &= (\sigma_d(\mathbf{Y} \boldsymbol{\mu}^*))^2 \\
    &\geq 
    (\sigma_d(\mathbf{Y})
    \sigma_d(\boldsymbol{\mu}^*))^2
    \\
    &= \lambda_d(\mathbf{Y}^T \mathbf{Y})
\lambda_d(\boldsymbol{\mu}^* \boldsymbol{\mu}^{*^T})
    \\
    &\geq \left( n_{(K-d+1)} \right)
    \frac{1}{\kappa_0} \lambda_1(\boldsymbol{\mu}^* \boldsymbol{\mu}^{*^T}) 
\end{aligned}
\end{equation*}
where $n_j=\sum_{i=1}^n \mathbb{I}(\tilde{\boldsymbol{\mu}}_i=\tilde{\boldsymbol{\mu}}^{(j)})$ denote the number of models with the $j$-th distinct population mean response, and $n_{(1)} \leq n_{(2)} \leq \dots \leq n_{(K)}$ denote the order statistics, and $\kappa_0=\frac{
\lambda_1(\boldsymbol{\mu}^* \boldsymbol{\mu}^{*^T})
}
{
\lambda_d(\boldsymbol{\mu}^* \boldsymbol{\mu}^{*^T})}$ is the condition number of the matrix $\boldsymbol{\mu}^* \boldsymbol{\mu}^{*^T}$.
\newline
Similarly, we can also prove
\begin{equation*}
    \lambda_1(\mathbf{B})
    \geq 
   \left( n_{(K)} \right)
   \frac{1}{\kappa_0}
\lambda_1(\boldsymbol{\mu}^* \boldsymbol{\mu}^{*^T}).
\end{equation*}
From the $2$nd condition, we can say that there exists a constant $C^{(d)}>0$ such that
\begin{equation*}
    \lambda_d(\mathbf{B}) \geq 
    C^{(d)}n
\end{equation*}
for all $n$ sufficiently large.
\newline
\newline
\subsection{Proofs of supporting propositions}
\label{Subsec:Proofs_of_supporting_propositions}
\begin{customproposition}{A.1} 
\label{Prop:R1_bound}
In our setting, suppose 
$
\frac{(2 L \omega C \sqrt{\Gamma})n}{\sqrt{mr}} +
\frac{\Gamma}{r}
\leq 
\frac{|\lambda_d|}{2}
$. Then,
\newline
$\text{with probability at least }  (1-2 e^{-\frac{n}{(C-1)^2}})$,
\begin{equation*}
    \left\lVert
\mathbf{R}_1
    \right\rVert
    \leq 
    \frac{\sqrt{2}}{\sqrt{|\lambda_d|}}
    \left(
\frac{(2 L \omega C \sqrt{\Gamma}) n}{\sqrt{m r}}+\frac{\Gamma}{r}
    \right)
    .
\end{equation*}
\end{customproposition}
\textbf{Proof.} 
Recall that $\mathbf{R}_1= -\mathbf{U} \mathbf{U}^T (\hat{\mathbf{B}}-\mathbf{B})
\hat{\mathbf{U}}
|
\hat{\boldsymbol{\Lambda}}
|^{-\frac{1}{2}}
$, which implies
$
\left\lVert
\mathbf{R}_1
\right\rVert
\leq 
\left\lVert
\hat{\mathbf{B}}-\mathbf{B}
\right\rVert
\left\lVert
|\hat{\boldsymbol{\Lambda}}|^{-\frac{1}{2}}
\right\rVert.
$
\newline
 Using  \Cref{Thm:Noise_CMDS_dissimilarity_bound} and Lemma \Cref{Lm:lemma_3}, 
\begin{equation*}
\begin{aligned}
    &\left\lVert
\hat{\mathbf{B}}-\mathbf{B}
    \right\rVert \leq 
    \frac{(2 L \omega C \sqrt{\Gamma}) n}{\sqrt{m r}}+\frac{\Gamma}{r}
    \text{ with probability at least }  (1-e^{-\frac{n}{(C-1)^2}})
    ,
    \text{ [from  \Cref{Thm:Noise_CMDS_dissimilarity_bound}]}
    \\ 
    &\left\lVert
|\hat{\boldsymbol{\Lambda}}|^{-\frac{1}{2}}
    \right\rVert
    \leq \frac{\sqrt{2}}{\sqrt{|\lambda_d|}}
    \text{ with probability at least } (1-e^{-\frac{n}{(C-1)^2}}),
    \text{ [from Lemma \Cref{Lm:lemma_3}]}
\end{aligned}
\end{equation*}
Thus, $\text{with probability at least }  (1-2e^{-\frac{n}{(C-1)^2}})$,
\begin{equation*}
\begin{aligned}
    \left\lVert
\mathbf{R}_1
    \right\rVert
    \leq 
    \frac{\sqrt{2}}{\sqrt{|\lambda_d|}}
    \left(
\frac{(2 L \omega C \sqrt{\Gamma}) n}{\sqrt{m r}}+\frac{\Gamma}{r}
\right).
\end{aligned}
\end{equation*}
\newline
\newline
\newline
\begin{customproposition}{A.2}
\label{Prop:R2_bound}
In our setting, 
suppose 
$
\frac{(2 L \omega C \sqrt{\Gamma}) n}{\sqrt{mr}} +
\frac{\Gamma}{r}
\leq 
\frac{|\lambda_d|}{2}
$. Then, 
\newline
$\text{with probability at least } (1-2 e^{-\frac{n}{(C-1)^2}})$,
\begin{equation*}
\begin{aligned}
   \left\lVert
\mathbf{R}_2
    \right\rVert
    \leq 
 \frac{\sqrt{2}}{1+\sqrt{2}}
    \frac{d}{\sqrt{|\lambda_d|}}
    \left(
\frac{(2 L \omega C \sqrt{\Gamma}) n}{\sqrt{m r}}+\frac{\Gamma}{r}
    \right)
.
\end{aligned}
\end{equation*}
\end{customproposition}
\textbf{Proof.} Recall that,
\begin{equation*}
\begin{aligned}
    \mathbf{R}_2=
    \mathbf{U} 
    (\mathbf{U}^T \hat{\mathbf{U}}
    |\hat{\boldsymbol{\Lambda}}|^{\frac{1}{2}} -
    |\boldsymbol{\Lambda}|^{\frac{1}{2}} 
    \mathbf{U}^T 
    \hat{\mathbf{U}}
    ) \implies
    \left\lVert
\mathbf{R}_2
    \right\rVert
    \leq 
    \left\lVert
\mathbf{U}^T 
\hat{\mathbf{U}} |\hat{\boldsymbol{\Lambda}}|^{\frac{1}{2}} - 
|\boldsymbol{\Lambda}|^{\frac{1}{2}} 
\mathbf{U}^T
\hat{\mathbf{U}}
    \right\rVert.
\end{aligned}
\end{equation*}
Using Lemma \ref{Lm:lemma_4}, we deduce that $\text{with probability at least }  (1-2e^{-\frac{n}{(C-1)^2}})$,  
\begin{equation*}
    \left\lVert
\mathbf{R}_2
    \right\rVert
    \leq 
    \frac{\sqrt{2}}{1+\sqrt{2}}
    \frac{d}{\sqrt{|\lambda_d|}}
    \left(
\frac{(2 L \omega C \sqrt{\Gamma}) n}{\sqrt{m r}}+\frac{\Gamma}{r}
    \right)
    .
\end{equation*}
\newline
\newline
\begin{customproposition}{A.3}
\label{Prop:R3_bound}
     In our setting,
suppose 
$
\frac{(2 L \omega C \sqrt{\Gamma}) n}{\sqrt{m r}} +
\frac{\Gamma}{r}
\leq 
\frac{|\lambda_d|}{2}
$. Then, 
$\text{with probability at least }  (1-e^{-\frac{n}{(C-1)^2}})$, 
\begin{equation*}
\begin{aligned}
    \left\lVert 
\mathbf{R}_3
    \right\rVert 
    \leq 
    \frac{4d}{|\lambda_d|^2}\sqrt{|\lambda_1|}
    \left(
\frac{(2 L \omega C \sqrt{\Gamma} ) n}{\sqrt{m r}}+\frac{\Gamma}{r}
    \right)^2
    .
\end{aligned}
\end{equation*}
\end{customproposition}
\textbf{Proof.} We know,
\begin{equation*} 
\mathbf{R}_3
 =
\mathbf{U}
|\boldsymbol{\Lambda}|^{\frac{1}{2}} (\mathbf{U}^T \hat{\mathbf{U}}- \mathbf{W}_*),
\end{equation*}
which gives us
\begin{equation*}
\begin{aligned}
\left\lVert 
\mathbf{R}_3
\right\rVert
&\leq
\left\lVert 
\mathbf{B}
\right\rVert^{\frac{1}{2}}
\left\lVert
\mathbf{U}^T \hat{\mathbf{U}}
- \mathbf{W}_* 
\right\rVert \\
&\leq
\left\lVert 
\mathbf{B}
\right\rVert^{\frac{1}{2}}
\left\lVert
\mathbf{U}^T \hat{\mathbf{U}} -
\mathbf{W}_*
\right\rVert_F
\\
&= 
\left\lVert
\mathbf{B}
\right\rVert^{\frac{1}{2}}
\left\lVert
\mathbf{I}- \mathrm{cos}\boldsymbol{\Theta}(\mathbf{U},\hat{\mathbf{U}})
\right\rVert_F \\
&\leq 
\left\lVert
\mathbf{B}
\right\rVert^{\frac{1}{2}}
\left\lVert
\mathrm{sin} \boldsymbol{\Theta}(\mathbf{U},\hat{\mathbf{U}})
\right\rVert_F^2 \\
&\leq
\frac{4d}{\lambda_d^2}
\left\lVert
\mathbf{B}
\right\rVert^{\frac{1}{2}}
\left\lVert 
\hat{\mathbf{B}}-\mathbf{B}
\right\rVert^2,
\end{aligned}
\end{equation*}
where the last inequality follows from Davis-Kahan Theorem
\citep{yu2015useful}. 
\newline 
Thus, $\text{with probability at least }  (1-e^{-\frac{n}{(C-1)^2}})$,
\begin{equation*}
\begin{aligned}
    \left\lVert 
\mathbf{R}_3
    \right\rVert
    \leq 
    \frac{4 d}{|\lambda_d|^2}
    \sqrt{|\lambda_1|}
    \left(
\frac{(2 L \omega C \sqrt{\Gamma}) n}{\sqrt{m r}}+\frac{\Gamma}{r}
    \right)^2.
\end{aligned}
\end{equation*}
\newline
\newline
\begin{customproposition}{A.4}
\label{Prop:R4_bound}
In our setting, 
suppose 
$
\frac{(2 L \omega C \sqrt{\Gamma}) n}{m r} + \frac{\Gamma}{r}
\leq \frac{|\lambda_d|}{2}
$. Then, $\text{with probability at least }  (1-e^{-\frac{n}{(C-1)^2}})$,
\begin{equation*}
\begin{aligned}
    \left\lVert 
\mathbf{R}_4
    \right\rVert
    \leq 
    \frac{1}{|\lambda_d|^{\frac{3}{2}}}
\left( 
\frac{(2 L \omega C \sqrt{\Gamma}) n}{\sqrt{m r}}+\frac{\Gamma}{r}
\right)^2.
\end{aligned}
\end{equation*}
\end{customproposition}
\textbf{Proof.} 
We know, 
\begin{equation*}
    \mathbf{R}_4 =
    (\hat{\mathbf{B}}-\mathbf{B})
    (\hat{\mathbf{U}} \hat{\mathbf{U}}^T \mathbf{U}-
    \mathbf{U})
    |\boldsymbol{\Lambda}|^{-\frac{1}{2}}. 
\end{equation*}
Hence, 
\begin{equation*}
\begin{aligned}
    \left\lVert 
\mathbf{R}_4
\right\rVert &\leq
\left\lVert 
\hat{\mathbf{B}}-\mathbf{B}
\right\rVert
\left\lVert 
\mathrm{sin} \boldsymbol{\Theta}
(\mathbf{U},\hat{\mathbf{U}})
\right\rVert
\frac{1}{\sqrt{|\lambda_d|}} \\
&\leq 
\frac{1}{|\lambda_d|^{\frac{3}{2}}}
\left\lVert 
\hat{\mathbf{B}}-\mathbf{B}
\right\rVert^2,
\end{aligned}
\end{equation*}
using Davis-Kahan theorem from \cite{chen2021spectral}. Using  \Cref{Thm:Noise_CMDS_dissimilarity_bound}, we get that 
\newline
$\text{with probability at least } (1-e^{-\frac{n}{(C-1)^2}})$,
\begin{equation*}
    \left\lVert
\mathbf{R}_4
\right\rVert
\leq 
\frac{1}{|\lambda_d|^{\frac{3}{2}}}
\left(
\frac{(2 L \omega C \sqrt{\Gamma}) n}{\sqrt{m r}}+\frac{\Gamma}{r}
\right)^2.
\end{equation*}
\newline
\newline
\begin{customproposition}{A.5}
\label{Prop:R5_bound}
In our setting, 
suppose 
$
\frac{(2 L \omega C \sqrt{\Gamma}) n}{\sqrt{m r}} +
\frac{\Gamma}{r}
\leq 
\frac{|\lambda_d|}{2}
$. Then, 
\newline
$\text{with probability at least } (1-4 e^{-\frac{n}{(C-1)^2}})$,
\begin{equation*}
\begin{aligned}
    \left\lVert 
\mathbf{R}_5
    \right\rVert 
    \leq   
\frac{2d}{(1+\sqrt{2})|\lambda_d|^{\frac{3}{2}}}
\left(
\frac{(2 L \omega C \sqrt{\Gamma}) n}{\sqrt{m r}}+\frac{\Gamma}{r}
\right)^2
.
\end{aligned}
\end{equation*}
\end{customproposition}
\textbf{Proof.} We know, 
\begin{equation*}
    \mathbf{R}_5=
    -(\hat{\mathbf{B}}-\mathbf{B})
    \hat{\mathbf{U}}
    (\hat{\mathbf{U}}^T \mathbf{U} |\boldsymbol{\Lambda}|^{-\frac{1}{2}} -    |\hat{\boldsymbol{\Lambda}}|^{-\frac{1}{2}}
    \hat{\mathbf{U}}^T \mathbf{U}
    )
\end{equation*}
which implies 
\begin{equation*}
\left\lVert 
\mathbf{R}_5
\right\rVert \leq 
\left\lVert 
\hat{\mathbf{B}}-\mathbf{B}
\right\rVert
\left\lVert
\hat{\mathbf{U}}^T \mathbf{U}
|\boldsymbol{\Lambda}|^{-\frac{1}{2}}-
|\hat{\boldsymbol{\Lambda}}|^{-\frac{1}{2}} \hat{\mathbf{U}}^T \mathbf{U}
\right\rVert
.
\end{equation*}
Now recall that 
\begin{equation*}
\begin{aligned}
    &\left\lVert 
\hat{\mathbf{B}}-\mathbf{B}
    \right\rVert
    \leq 
    \left(
\frac{(2 L \omega C \sqrt{\Gamma}) n}{\sqrt{ m r}} + 
\frac{\Gamma}{r}
    \right)
    \text{ with probability at least } (1-e^{-\frac{n}{(C-1)^2}}) \text{ [from  \Cref{Thm:Noise_CMDS_dissimilarity_bound}]}, \\
    &\left\lVert 
\hat{\mathbf{U}}^T 
\mathbf{U} 
|\boldsymbol{\Lambda}|^{-\frac{1}{2}} -
|\hat{\boldsymbol{\Lambda}}|^{-\frac{1}{2}} 
\hat{\mathbf{U}}^T
\mathbf{U}
    \right\rVert
    \leq 
    \frac{2d}{(1+\sqrt{2}) |\lambda_d|^{\frac{3}{2}}}
    \left(
\frac{(2 L \omega C \sqrt{\Gamma}) n}{\sqrt{m r}} +
\frac{\Gamma}{r}
    \right)
    \text{ with probability at least } (1-3 e^{-\frac{n}{(C-1)^2}})
    \text{ [from Lemma \ref{Lm:lemma_5}]}.
\end{aligned}
\end{equation*}
Thus, using Bonferroni's Inequality,
$\text{with probability at least }  (1-4 e^{-\frac{n}{(C-1)^2}})$,
\begin{equation*}
\begin{aligned}
   \left\lVert 
\mathbf{R}_5
   \right\rVert
   \leq 
   \frac{2d}{(1+\sqrt{2})|\lambda_d|^{\frac{3}{2}}}
   \left(
\frac{(2 L \omega C \sqrt{\Gamma})n}{\sqrt{m r}} + 
\frac{\Gamma}{r}
   \right)^2.
\end{aligned}
\end{equation*}
\newline
\newline
\begin{customproposition}{A.6}
\label{Prop:R6_bound}
In our setting, 
suppose 
$
\frac{(2 L \omega C \sqrt{\Gamma}) n}{\sqrt{m r}} + 
\frac{\Gamma}{r}
\leq 
\frac{|\lambda_d|}{2}
$. Then,
\newline
$\text{with probability at least } (1-2 e^{-\frac{n}{(C-1)^2}})$,
\begin{equation*}
\begin{aligned}
    \left\lVert 
\mathbf{R}_6
    \right\rVert
    \leq 
\frac{4 \sqrt{2} d}{|\lambda_d|^{\frac{5}{2}}}
\left(
\frac{(2 L \omega C \sqrt{\Gamma}) n}{\sqrt{m r}}+\frac{\Gamma}{r}
\right)^{3}
    .
\end{aligned}
\end{equation*}
\end{customproposition}
\textbf{Proof.} 
We know,
$
\mathbf{R}_6=
(\hat{\mathbf{B}}-\mathbf{B})
\hat{\mathbf{U}}
|\hat{\boldsymbol{\Lambda}}|^{-\frac{1}{2}}
\mathbf{I}_{p}
(\mathbf{W}_*^T-
\hat{\mathbf{U}}^T \mathbf{U})
$.
Using the \textit{proof of Lemma C.4 in Section C.1} (\cite{agterberg2022joint}), 
\begin{equation*}
\begin{aligned}
    \left\lVert
\mathbf{W}_* -
\mathbf{U}^T \hat{\mathbf{U}}
    \right\rVert
    &\leq 
    \left\lVert
\mathbf{I}- \mathrm{cos} \boldsymbol{\Theta}
    \right\rVert \\
    &\leq 
    \left\lVert
\mathbf{I}-\mathrm{cos} \boldsymbol{\Theta}
    \right\rVert_F \\
    &\leq 
    \left\lVert
\mathrm{sin} \boldsymbol{\Theta}
(\mathbf{U},\hat{\mathbf{U}})
    \right\rVert_F^2 \\
    &\leq 
    \frac{4d}{\lambda_d^2}
    \left\lVert
\hat{\mathbf{B}}-\mathbf{B}
    \right\rVert^2. 
\end{aligned}
\end{equation*}
We know,  
\begin{equation*}
\begin{aligned}
    &\left\lVert
\hat{\mathbf{B}}-\mathbf{B}
    \right\rVert
    \leq 
    \left(
\frac{(2 L \omega C \sqrt{\Gamma}) n}{\sqrt{m r}}+\frac{\Gamma}{r}
    \right)
    \text{ with probability at least } (1-e^{-\frac{n}{(C-1)^2}})
    \text{ [from \textit{Theorem \ref{Thm:Noise_CMDS_dissimilarity_bound}}]}
    , \\
    &\left\lVert
|\hat{\boldsymbol{\Lambda}}|^{-\frac{1}{2}}
    \right\rVert
    \leq \frac{\sqrt{2}}{\sqrt{|\lambda_d|}}
    \text{ with probability at least } (1-e^{-\frac{n}{(C-1)^2}})
    \text{ [from \textit{Lemma \ref{Lm:lemma_3}}]}.
\end{aligned}
\end{equation*}
Thus, using Bonferroni's Inequality,  $\text{with probability at least } (1-2 e^{-\frac{n}{(C-1)^2}})$,
\begin{equation*}
    \left\lVert
\mathbf{R}_6
    \right\rVert
    \leq 
\frac{4 \sqrt{2} d}{|\lambda_d|^{\frac{5}{2}}}
\left(
\frac{(2 L \omega C \sqrt{\Gamma}) n}{\sqrt{m r}} +
\frac{\Gamma}{r}
\right)^{3}.
\end{equation*}
\subsection{Proofs of Lemmas}
\label{Subsec:Proofs_of_lemmas}
\textit{
\begin{customlemma}{A.1}
\label{Lm:lemma_2}
In our setting, suppose \textit{Assumption \ref{Asm:Response_distribution}} holds, and $\frac{(2 L \omega C \sqrt{\Gamma})n}{\sqrt{mr}}+\frac{\Gamma}{r} \leq \frac{|\lambda_d|}{2}$. Additionally,
, define $\mathbf{H} \in \mathbb{R}^{d \times d}$
such that for all $k,l \in [d]$,
\begin{equation*}
\mathbf{H}_{kl}=
\frac{1}{
|\lambda_l|^{\frac{1}{2}}+
|\hat{\lambda}_k|^{\frac{1}{2}}
}. 
\end{equation*}
Then, $\text{with probability at least }  (1-e^{-\frac{n}{(C-1)^2}})$, 
\begin{equation*}
    \left\lVert
\mathbf{H}
    \right\rVert
    \leq
    \left(
    \frac{\sqrt{2}}{1+\sqrt{2}}
    \right)
    \frac{d}{\sqrt{|\lambda_d|}}.
\end{equation*}
\end{customlemma}
}
\textbf{Proof.}
Using Triangle Inequality and Weyl's Inequality, 
\begin{equation*}
    |\lambda_k|- 
    |\hat{\lambda}_k|
    \leq |\lambda_k-\hat{\lambda}_k| \leq
    \left\lVert
\mathbf{B}-\hat{\mathbf{B}}
    \right\rVert.
\end{equation*}
Thus, 
$
|\hat{\lambda}_k|
\geq |\lambda_k|-
\left\lVert
\mathbf{B}-\hat{\mathbf{B}}
\right\rVert
$. 
\newline
Note that 
$
\left\lVert
\hat{\mathbf{B}}-\mathbf{B}
\right\rVert \leq 
\frac{(2 L \omega C \sqrt{\Gamma}) n}{\sqrt{mr}}
+ \frac{\Gamma}{r}
<\frac{|\lambda_d|}{2}
$ with $\text{with probability at least }  \left( 1-e^{-\frac{n}{(C-1)^2}} \right)$.
\newline
Thus, 
$
|\hat{\lambda}_k|
\geq \frac{|\lambda_d|}{2}
$ $\text{with probability at least } \left( 1-e^{-\frac{n}{(C-1)^2}} \right)$. 
\newline
Thus,
$\text{with probability at least } \left( 1-e^{-\frac{n}{(C-1)^2}} \right)$,
\begin{equation*}
\begin{aligned}
    \mathbf{H}_{kl}
    &=\frac{1}{|\lambda_l|^{\frac{1}{2}}+|\hat{\lambda}_k|^{\frac{1}{2}}} \\
    &\leq 
    \frac
    {1}
    {
    |\lambda_d|^{\frac{1}{2}}+
    \sqrt{\frac{|\lambda_d|}{2}}
    } \text{ for all $k,l \in [d]$} \\
    &\implies
    \left\lVert
\mathbf{H}
    \right\rVert
    \leq
    \left(
    \frac{\sqrt{2}}{1+\sqrt{2}}
    \right)
    \frac{d}{\sqrt{|\lambda_d|}}.
\end{aligned}
\end{equation*}
\newline
\newline
\newline
\textit{
\begin{customlemma}{A.2}
\label{Lm:lemma_3}
In our setting, suppose \textit{Assumption \ref{Asm:Response_distribution}} holds, and
suppose 
$
\frac{(2 L \omega C \sqrt{\Gamma})n}{\sqrt{mr}}+
\frac{\Gamma}{r}
\leq \frac{|\lambda_d|}{2}
$. Then, $\text{with probability at least } \left( 1-e^{-\frac{n}{(C-1)^2}} \right)$,
\begin{equation*}
    \left\lVert
\hat{\boldsymbol{\Lambda}}^{-1}
    \right\rVert \leq 
    \frac{2}{|\lambda_d|}.
\end{equation*}
\end{customlemma}
}
\textbf{Proof.} Note that 
$\hat{\boldsymbol{\Lambda}}^{-1}=\mathrm{diag}(\frac{1}{\hat{\lambda}_1},\dots \frac{1}{\hat{\lambda}_d})$. We know (see proof of \textit{Lemma 2}), under given conditions, for every $k \in [d]$, 
$|\hat{\lambda}_k| \geq \frac{|\lambda_d|}{2}$ with $\text{with probability at least } \left( 1-e^{-\frac{n}{(C-1)^2}} \right)$. Thus,  $\text{with probability at least } \left( 1-e^{-\frac{n}{(C-1)^2}} \right)$,
\begin{equation*}
    \left\lVert
\hat{\boldsymbol{\Lambda}}^{-1}
    \right\rVert 
    =
    \max_{k \in [d]} \frac{1}{|\hat{\lambda}_k|}
    \leq 
    \frac{2}{|\lambda_d|}.
\end{equation*}
\newline
\newline
\newline
\begin{customlemma}{A.3}
\label{Lm:lemma_4}
In our setting, suppose 
\textit{Assumption \ref{Asm:Response_distribution}} holds, and
suppose 
$
\frac{(2 L \omega C\sqrt{\Gamma})n}{\sqrt{mr}}
+ \frac{\Gamma}{r}
\leq 
\frac{|\lambda_d|}{2}
$. Then, $\text{with probability at least } \left( 1-2 e^{-\frac{n}{(C-1)^2}} \right)$, 
\begin{equation*}
\left\lVert
\hat{\mathbf{U}}^T \mathbf{U}
|\boldsymbol{\Lambda}|^{\frac{1}{2}}- 
|\hat{\boldsymbol{\Lambda}}|^{\frac{1}{2}} \hat{\mathbf{U}}^T \mathbf{U}
\right\rVert
\leq 
\frac{\sqrt{2}}{1+\sqrt{2}}
\frac{ d}{\sqrt{|\lambda_d|}}
\left(
\frac{(2 L \omega C \sqrt{\Gamma}) n}{\sqrt{m r}}+\frac{\Gamma}{r}
\right)
.
\end{equation*}
\end{customlemma}
\textbf{Proof.} Following \textit{Section C.1} in \cite{agterberg2022joint}, 
\begin{equation*}
\begin{aligned}
    \left\lVert
\hat{\mathbf{U}}^T \mathbf{U}
|\boldsymbol{\Lambda}|^{\frac{1}{2}} - 
|\hat{\boldsymbol{\Lambda}}|^{\frac{1}{2}} \hat{\mathbf{U}}^T \mathbf{U}
    \right\rVert &=
    \left\lVert
\hat{\mathbf{U}}^T \mathbf{U}
|\boldsymbol{\Lambda}|^{\frac{1}{2}}  - 
|\hat{\boldsymbol{\Lambda}}|^{\frac{1}{2}} \hat{\mathbf{U}}^T \mathbf{U} 
    \right\rVert \\
    &=
    \left\lVert
\mathbf{H} \circ 
\left\lbrace
\hat{\mathbf{U}}^T (\hat{\mathbf{B}}-\mathbf{B})
\mathbf{U}
\right\rbrace
    \right\rVert \\
    &\leq 
\left\lVert
\mathbf{H}
\right\rVert
\left\lbrace
\left\lVert
\hat{\mathbf{U}}^T (\hat{\mathbf{B}}-\mathbf{B})
\mathbf{U}
\right\rVert 
\right\rbrace.
\end{aligned}
\end{equation*}
Recall that, under given conditions, 
\begin{equation*}
\begin{aligned}
    &\left\lVert
\mathbf{H}
    \right\rVert
    \leq 
    \left(
\frac{\sqrt{2}}{1+\sqrt{2}}
    \right)
    \frac{d}{\sqrt{|\lambda_d|}},
    \text{ with probability at least $\left( 1-e^{-\frac{n}{(C-1)^2}} \right)$} 
    \text{ [from Lemma \ref{Lm:lemma_2}]}
    \\
    &\left\lVert
\hat{\mathbf{U}}^T (\hat{\mathbf{B}}-\mathbf{B})
\mathbf{U}
    \right\rVert \leq 
\left\lVert 
\hat{\mathbf{B}}-\mathbf{B}
\right\rVert
\leq 
    \left(
\frac{(2 L \omega C \sqrt{\Gamma}) n}{\sqrt{m r}}+\frac{\Gamma}{r}
    \right),
    \text{ with probability at least  $\left( 1-e^{-\frac{n}{(C-1)^2}} \right)$}
    \text{ [from  \Cref{Thm:Noise_CMDS_dissimilarity_bound}]}
    .
\end{aligned}
\end{equation*}
\newline
Thus, combining the above two inequalities,
$\text{with probability at least } \left( 1-2e^{-\frac{n}{(C-1)^2}} \right)$
\begin{equation*}
\begin{aligned}
\left\lVert
\hat{\mathbf{U}}^T \mathbf{U} 
|\boldsymbol{\Lambda}|^{\frac{1}{2}} -
|\hat{\boldsymbol{\Lambda}}|^{\frac{1}{2}} 
\hat{\mathbf{U}}^T \mathbf{U}
\right\rVert
\leq 
\frac{\sqrt{2}}{1+\sqrt{2}}
\frac{ d}{\sqrt{|\lambda_d|}}
\left(
\frac{(2 L \omega C \sqrt{\Gamma}) n}{\sqrt{m r}}+\frac{\Gamma}{r}
\right)
.
\end{aligned}
\end{equation*}
\newline
\begin{customlemma}{A.4}
\label{Lm:lemma_5}
In our setting,
suppose 
\textit{Assumption \ref{Asm:Response_distribution}} holds, and
$
\frac{(2 L \omega C \sqrt{\Gamma})n}{\sqrt{mr}}
+ \frac{\Gamma}{r}
\leq 
\frac{|\lambda_d|}{2}
$. Then, $\text{with probability at least } \left( 1-3 e^{-\frac{n}{(C-1)^2}} \right)$, 
\begin{equation*}
    \left\lVert
\hat{\mathbf{U}}^T \mathbf{U}
|\boldsymbol{\Lambda}|^{-\frac{1}{2}} 
-
|\hat{\boldsymbol{\Lambda}}|^{-\frac{1}{2}}
\hat{\mathbf{U}}^T \mathbf{U}
    \right\rVert
    \leq 
    \frac{2d}{(1+\sqrt{2})|\lambda_d|^{\frac{3}{2}}}
    \left(
\frac{(2 L \omega C \sqrt{\Gamma}) n}{\sqrt{m r}}+\frac{\Gamma}{r}
    \right)
    .
\end{equation*}
\end{customlemma}
\textbf{Proof.} 
First note that (from proof of \textit{Lemma C.4} in \cite{agterberg2022joint}),
\begin{equation*}
\begin{aligned}
\left\lVert
\hat{\mathbf{U}}^T \mathbf{U}
|\boldsymbol{\Lambda}|^{-\frac{1}{2}} 
-
|\hat{\boldsymbol{\Lambda}}|^{-\frac{1}{2}}
\hat{\mathbf{U}}^T \mathbf{U}
\right\rVert &=
\left\lVert
|\hat{\boldsymbol{\Lambda}}|^{-\frac{1}{2}}
\left(
|\hat{\boldsymbol{\Lambda}}|^{\frac{1}{2}}
\hat{\mathbf{U}}^T \mathbf{U}
-
\hat{\mathbf{U}}^T \mathbf{U}
|\boldsymbol{\Lambda}|^{\frac{1}{2}}
\right)
|\boldsymbol{\Lambda}|^{-\frac{1}{2}}
\right\rVert.
\end{aligned}
\end{equation*}
Note that,
\begin{equation*}
\begin{aligned}
    &\left\lVert 
|\hat{\boldsymbol{\Lambda}}|^{-\frac{1}{2}}
    \right\rVert
    \leq 
    \frac{\sqrt{2}}{\sqrt{|\lambda_d|}}, \text{with probability at least $\left( 1-e^{-\frac{n}{(C-1)^2}} \right)$}
    \text{ [from \ref{Lm:lemma_3}]}
    \\
 &\left\lVert   
|\hat{\boldsymbol{\Lambda}}|^{\frac{1}{2}}
\hat{\mathbf{U}}^T 
\mathbf{U} -
\hat{\mathbf{U}}^T
\mathbf{U}
|\boldsymbol{\Lambda}|^{\frac{1}{2}}
    \right\rVert
    \leq 
    \frac{\sqrt{2}}{1+\sqrt{2}}
    \frac{d}{\sqrt{|\lambda_d|}}
    \left(
\frac{(2 L \omega C \sqrt{\Gamma}) n}{\sqrt{m r}} +
\frac{\Gamma}{r}
    \right) 
    \text{with probability at least $\left( 1-2e^{-\frac{n}{(C-1)^2}} \right)$}
    \text{ [from  \Cref{Lm:lemma_4}]}.
\end{aligned}
\end{equation*}
Hence, using Bonferroni's Inequality,
$\text{with probability at least } \left( 1-3e^{-\frac{n}{(C-1)^2}} \right)$,
\begin{equation*}
\begin{aligned}
    \left\lVert
\hat{\mathbf{U}}^T
\mathbf{U}
|\boldsymbol{\Lambda}|^{-\frac{1}{2}} -
|\hat{\boldsymbol{\Lambda}}|^{-\frac{1}{2}}
\hat{\mathbf{U}}^T
\mathbf{U}
    \right\rVert
    \leq 
    \frac{2d}{(1+\sqrt{2})|\lambda_d|^{\frac{3}{2}}}
    \left(
\frac{(2 L \omega C \sqrt{\Gamma}) n}{\sqrt{ m r}}+
\frac{\Gamma}{r}
    \right)
\end{aligned}
\end{equation*}
\subsection{Key results used from other papers}
\label{Subsec:Results_from_others}
\begin{customthm}{A} (\cite{agterberg2022joint})
\label{Thm:ref_Jagt_thm}
In our setting, 
\begin{equation*}
\begin{aligned}
    \hat{\boldsymbol{\psi}} \mathbf{W}_*^T-\boldsymbol{\psi}=
    (\hat{\mathbf{B}}-\mathbf{B})
    \mathbf{U}|\boldsymbol{\Lambda}|^{-\frac{1}{2}} +
    \mathbf{R}_1 \mathbf{W}_*^T+
    \mathbf{R}_2 \mathbf{W}_*^T+
    \mathbf{R}_3 \mathbf{W}_*^T+
    \mathbf{R}_4+
    \mathbf{R}_5+
    \mathbf{R}_6
\end{aligned}
\end{equation*}
where
\begin{equation*}
\begin{aligned}
    &\mathbf{R}_1=-\mathbf{U} \mathbf{U}^T (\hat{\mathbf{B}}-\mathbf{B})
    \hat{\mathbf{U}}
    |\hat{\boldsymbol{\Lambda}}|^{-\frac{1}{2}}, \\
    &\mathbf{R}_2= 
    \mathbf{U}
    (\mathbf{U}^T \hat{\mathbf{U}} |\hat{\boldsymbol{\Lambda}}|^{\frac{1}{2}}-|\boldsymbol{\Lambda}|^{\frac{1}{2}}\mathbf{U}^T \hat{\mathbf{U}}), \\
    &\mathbf{R}_3= \mathbf{U}|\boldsymbol{\Lambda}|^{\frac{1}{2}}(\mathbf{U}^T \hat{\mathbf{U}}-\mathbf{W}_*),\\
    &\mathbf{R}_4= (\hat{\mathbf{B}}-\mathbf{B})
    (\hat{\mathbf{U}} \hat{\mathbf{U}}^T \mathbf{U}-\mathbf{U})|\boldsymbol{\Lambda}|^{-\frac{1}{2}}, \\
    &\mathbf{R}_5=-(\hat{\mathbf{B}}-\mathbf{B})
    \hat{\mathbf{U}}
    (\hat{\mathbf{U}}^T \mathbf{U} |\boldsymbol{\Lambda}|^{-\frac{1}{2}}-
    |\hat{\boldsymbol{\Lambda}}|^{-\frac{1}{2}} 
    \hat{\mathbf{U}}^T \mathbf{U} ), \\
    &\mathbf{R}_6=
    (\hat{\mathbf{B}}-\mathbf{B})
    \hat{\mathbf{U}}|\hat{\boldsymbol{\Lambda}}|^{-\frac{1}{2}}
    (\mathbf{W}_*^T-
    \hat{\mathbf{U}}^T \mathbf{U}).
\end{aligned}
\end{equation*}
\end{customthm}
\end{document}